\documentclass{article}

\usepackage[dblblindworkshop, final]{neurips_2025}

\usepackage[utf8]{inputenc} 
\usepackage[T1]{fontenc}    
\usepackage{hyperref}       
\usepackage{url}            
\usepackage{booktabs}       
\usepackage{amsfonts}       
\usepackage{nicefrac}       
\usepackage{microtype}      
\usepackage{xcolor}         

\usepackage{wrapfig}

\usepackage{amsmath,amsthm,amssymb,amsfonts,mathtools} 
\usepackage{bbm, bm}

\usepackage{graphicx} 
\usepackage{subcaption}

\title{Conditional Denoising Diffusion Autoencoders for Wireless Semantic Communications}
\workshoptitle{AI and ML for Next-Generation Wireless Communications and Networking (AI4NextG @ NeurIPS’25)}

\author{%
Mehdi Letafati \quad Samad Ali \quad Matti Latva-aho  \\
Centre for Wireless Communications\\ University of Oulu, Finland \\
\texttt{\{mehdi.letafati,samad.ali,matti.latva-aho\}@oulu.fi}\\
}

\newcommand{\E}{\mathbb{E}}
\newcommand{\vect}[1]{\mathbf{#1}}
\newcommand{\ysem}{\vect{y}_{\mathsf{sem}}}

\newcommand{\xt}{\vect{x}_t}
\newcommand{\xtone}{\vect{x}_{t-1}}
 
\newcommand{\norm}[1]{\left\lVert#1\right\rVert}

\DeclareMathOperator*{\argmin}{arg\,min}

\makeatletter 
\newcommand{\subalign}[1]{%
  \vcenter{%
    \Let@ \restore@math@cr \default@tag
    \baselineskip\fontdimen10 \scriptfont\tw@
    \advance\baselineskip\fontdimen12 \scriptfont\tw@
    \lineskip\thr@@\fontdimen8 \scriptfont\thr@@
    \lineskiplimit\lineskip
    \ialign{\hfil$\m@th\scriptstyle##$&$\m@th\scriptstyle{}##$\hfil\crcr
      #1\crcr
    }%
  }%
}

\newtheorem{theorem}{Theorem}
\newtheorem{lemma}{Lemma}

\begin{document}

\maketitle
\begin{abstract}
Semantic communication  (SemCom) systems aim to learn the mapping from low-dimensional semantics to high-dimensional ground-truth.  
While this is more akin to a ``domain translation'' problem,  
existing frameworks
typically emphasize on channel-adaptive neural encoding-decoding schemes, lacking full exploration of signal distribution. Moreover, such methods so far have employed autoencoder-based  architectures, where the encoding is tightly coupled to a matched decoder, causing \emph{scalability issues} in practice.  To address these gaps, diffusion autoencoder models are proposed for wireless SemCom. The goal is to learn a ``semantic-to-clean'' mapping, from the semantic space to the ground-truth probability distribution.  
A neural encoder at semantic transmitter extracts the high-level semantics, and a conditional diffusion model (CDiff) at the semantic receiver exploits the source distribution for signal-space denoising, while the received semantic latents are incorporated as the conditioning input to  ``steer'' the decoding process towards the  semantics intended by the transmitter.  
It is    
analytically proved that the proposed decoder model
is a \emph{consistent estimator} of the ground-truth data.
Furthermore, extensive simulations over CIFAR-10 and MNIST datasets are provided along with design insights, highlighting  the performance compared to legacy autoencoders and variational autoencoders (VAE). 
Simulations are further extended to the multi-user SemCom, identifying the dominating factors 
in a more realistic setup. 
\end{abstract}

\section{Introduction} 
Semantic communications (SemCom) has been considered as one of the enablers for a truly AI-native communication framework for next-generations (nextG) of 
wireless systems \cite{Beyond_bits}.  
In simple terms,  SemCom paradigm provides means to convey the most relevant information, where bits
are no longer the common currency between the application and physical layers.     
Recent advances in generative artificial intelligence (GenAI)  and its   capabilities in generating high-fidelity samples have boosted the interest in exploring its potentials for nextG systems  \cite{GenAI_PHY, generative_JSCC}.    
The evolution of  diffusion models has contributed to  the recent breakthroughs in non-language GenAI, with renowned solutions such as DALL.E 2 by OpenAI and Imagen by Google. 
Interestingly, recent studies have proven their applications in  wireless AI as well \cite{mit}--\cite{CDiff_TMLCN}. 
Nevertheless, diffusion models lack any mechanism to extract semantic meaning. 
Our motivation in this paper is to  smartly combine the capabilities of autoencoders in  extracting  decodable  latents, with the capabilities of diffusion models in discovering the signal space distributions, to come up with an enhanced SemCom framework for nextG wireless.

\subsection{Related Works}   
Autoencoder-based architectures under the framework of deep joint source-channel coding (deep-JSCC), have been widely adopted  for SemCom \cite{DeepJSCC_for_SemCom}. 
Semantic image transmission using diffusion models was proposed in \cite{High_perceptual_Image_DM} and \cite{INN}. A denoising diffusion probabilistic model (DDPM) was proposed to be appended to the deep-JSCC framework. 
However, this assumes  a very simplistic view of generative models, in the sense that it simply considers generative models as an \emph{additional module} to be added to a deep-JSCC scheme, only for further refinement, at the obvious cost of additional computation by adding one more AI processing block. 
Such simplistic view of using GenAI is not able to, for example, utilize generative modeling for replacing some of the legacy functionalities in deep-JSCC/SemCom framework. Moreover, the paper  does not exploit the full capabilities of generative models, such as their conditioning  mechanisms which can flexibly  incorporate additional contextual information to guide the models in reconstruction with desired fidelity.   
Diffusion models as ``channel denoisers'' were studied in \cite{CDDM} and \cite{CDDM_MUMIMO} 
by pre-pending the diffusion model to the deep-JSCC decoder.  
Nevertheless, such approach does not seem to be exploiting the full potential of diffusion model in learning complicated probability distributions, downgrading it to a simple channel denoiser.     
Furthermore, 
the it relies on the channel state information (CSI) knowledge at the  receiver. 
This seems to be conceptually in contradiction with the purpose of channel denoising, which is supposed to learn the CSI distributions, not receiving the CSIs directly.  
\vspace{-1mm}

\subsection{Contributions}  
\vspace{-1mm}   
Deep-JSCC-based SemCom frameworks with autoencoder  architectures as backbone mainly focus on channel-adaptive encoding and decoding, lacking  a powerful model for effectively learning signal distributions. Moreover, the decoder architecture  is tightly dependent to the  semantic encoder, which can cause \emph{scalability issues} in  practical deployments. Furthermore,  autoencoders seem to lack sample quality in  learning complicated distributions, lacking high-quality reconstruction performance.   
In this work, we propose diffusion autoencoder models for wireless SemCom.   
The idea is to use a neural encoder for extracting the high-level semantics, and a  
 conditional diffusion model (CDiff) as the decoder to explore the source data distribution and reconstruct samples,  while being guided by the semantic latents. In our scheme, the noisy semantics are incorporated as the conditioning input to the CDiff decoder.   
This way, semantic latents are distilled into the decoding process, which help ``steer'' the decoding process towards the  semantics intended by the transmitter. 
 For our CDiff decoder, we employ conditional DDPM, while the encoder is arbitrarily designed via prevalent convolution neural network (CNN)-based neural encoders.   
Our scheme 
offers advantages  over  generative models like variational autoencoders (VAE). 
While our scheme does not impose any constraint 
on the format of the semantic latent and allows the neural encoder to obtain the semantics as expressive as it can,  VAEs typically cast a standard normal distribution on the prior distribution of the latent space, that can compromise the decodability quality of semantics.   

With our diffusion-based decoder model at the receiver, the requirement to match the neural encoder and decoder pairs is dropped --- we do not employ the matched decoder counterpart of the neural encoder.  The reason is that the built-in conditioning mechanism of diffusion models is able to flexibly handle variable-length vectors corresponding to the semantic latents of different sizes, thanks to  padding \& masking conditioning mechanism.   
Thus, we not only enhance the semantic decoding performance, but  also mitigate the \emph{scalability issues} of existing deep-JSCC-based SemCom frameworks, saving computation and radio resources.     
Moreover, 
our decoder model works with a range of channel bandwidth ratios (CBR), and the CDiff decoder does not need to participate in end-to-end joint training with every single neural encoder. Once trained with an arbitrary encoder, the model is inherently capable of adapting to different neural encoders of different sizes through the diffusion  conditioning.  
Our visions is that receivers will have a general-purpose semantic decoder,  acting  as  a ``foundation'' decoder model in the future nextG systems.

To summarize, our contributions can be outlined as follows: i)   We present a novel machine learning model for SemCom systems, which is based on conditional denoising  diffusion models.  We present the data pipelines for training and inference, as well as the sampling framework.   ii)  While generative models are  \emph{probabilist  models} by nature,  we provide formal guarantees, using probability theory, to statistically  prove  that the proposed decoder model is a \emph{consistent estimator} of the ground-truth data. 
iii) We provide extensive simulation results over two different datasets, studying the effect of CBR, signal-to-noise ratio (SNR), and  multi-user semantic interference.
We highlight $55\%$ improvement in terms of 
the learned perceptual image patch similarity (LPIPS)  and $30\%$ improvement in terms of  structural similarity index measure (SSIM) compared to autoencoders  with matched encoder-decoder architecture, as well as  $50\%$ improvement in LPIPS compared to VAE benchmark.

\section{Problem Formulation}  
\label{gen_inst}

\vspace{-2mm}
Consider a  SemCom  system for data transmission  from a source node equipped with a semantic encoder  to a destination node equipped with a semantic decoder.  Source data is denoted by $\vect{x}_0\in \mathbb{R}^n$, following a non-trivial probability distribution ${\vect{x}}_0 \sim p_{\vect{x}_0}(\vect{x}_0)$. 
The semantic encoder, denoted by $\mathcal{E}(\cdot)$, 
is supposed to extract the high-level semantics of the source data, mapped to a semantic latent vector $\vect{x}_{\mathsf{sem}}$, i.e.,  
  $  \vect{x}_{\mathsf{sem}} = \mathcal{E}(\vect{x}_0; \bf{\phi})$,  
where the  encoder is parameterized by $\bf{\phi}$.  
Considering $\vect{x}_{\mathsf{sem}} \in \mathbb{R}^{2k}$, 
the channel input is treated 
as the equivalent $k$-dimensional complex-valued vector $\bm{z} \in \mathbb{C}^k$,  
normalized to average transmit power constraint $P$, i.e., $\quad  \frac{1}{k}\sum_{i=1}^k \mid z_i \mid^2 \leq P$.
The source data dimension, $n$, is referred to as the  {source bandwidth}, while the channel dimension $k$  characterizes  the {channel bandwidth}, and the  bandwidth compression ratio (BCR) is  defined as $k/n$ with  $k<n$.  This  can be seen as an indicator on the   spectral or temporal radio resources available \cite{DeepJSCC_for_SemCom}.  Communication channel is  modeled as a transfer function $\eta(\cdot)$, where   $ \eta(\bm{z}) =  \bm{z} + \bm{n}$, with $\bm{n} \sim \mathcal{CN}(\bm{0}, \sigma^2 \bm{I}_{k})$. 
Accordingly,  a noisy distorted version  of the semantic latents, denoted by $\vect{y}_{\mathsf{sem}}$,  is received by the semantic decoder $\mathcal{D}(\cdot)$ parametrized by $\bf{\theta}$, which  semantically reconstructs the data, using the received noisy semantic latent vectors, i.e., 
    $\hat{{\vect{x}}}_{\theta} = \mathcal{D}({\vect{y}}_{\sf{sem}}; \bf{\theta})$. 

The problem is to obtain at the semantic decoder, high-quality semantic reconstructions  $\hat{\vect{x}}\sim p_{\hat{\vect{x}}}(\hat{\vect{x}})$ close to the transmitter's  ground-truth data $\vect{x}_0\sim p_{\vect{x}_0}$ in the same signal space. 
The quality of the reconstruction is typically evaluated via distortion measures such as peak signal-to-noise ratio (PSNR).  
Nevertheless, 
the focus in SemCom frameworks is more on the perceptual reconstructions.  This can be achieved by utilizing  statistical distance measures (such as Kullback-Leibler (KL) divergence or f-divergence) between the generated probability distribution and the source probability  distribution. 
 Hence, the general objective function can be expressed as a weighted sum of the two objective functions, i.e., the distortion function $d_1(\vect{x}, \hat{\vect{x}})$ over the signal space, and the statistical divergence measures $d_2(P_X, P_{\hat{X}})$ over the probability distributions:  
\begin{align} 
\underset{\phi, \theta}{\text{minimize   }} 
     &\mathbb{E}_{{\vect{x}_0}\sim p_{{\vect{x}_0}}}  
     \mathbb{E}_{\hat{\vect{x}}\sim p_{\hat{\vect{x}}\vert\vect{x}_0}} 
     \left[ 
     d(\vect{x}_0, \hat{\vect{x}}_\theta)
     \right]  
      =  
     \lambda_1 
     \mathbb{E}_{p(\vect{x}_0, \hat{\vect{x}}_\theta)} 
     \left[ 
     d_1(\vect{x}_0, \hat{\vect{x}}_\theta)
     \right]
     +
     \lambda_2 d_2(p_{\vect{x}_0}, p_{\hat{\vect{x}}}).  
\end{align}

Solving this problem  can be viewed as a domain translation type of problem, 
where  the goal is to learn the semantic-to-clean mapping from noisy semantics to  ground-truths.  
Semantic data transmission can be treated  as a forward degradation process, which is typically assumed to be highly non-linear.  
Then the problem  can be viewed as the \emph{inverse problem,}  with the  received  semantics treated  as  ``measurements'' which are gone through a typically non-invertible forward process. This can be approximated by a forward operator $\mathcal{F}(\cdot)$, where $\vect{y}_{\sf{sem}} = \mathcal{F}(\vect{x}_0) \approx  \eta(\mathcal{E}({\vect{x}_0}; \phi))$. Then the problem is to invert the forward process and  semantically reconstruct the source data from the ``observations'', which can be realized by minimizing the following approximated loss function    
\begin{align}
    \mathcal{L}(\phi, \theta) :=  
\mathbb{E}  \left[   d\left({\vect{x}_0}, \mathcal{D}\left(\eta\left(\mathcal{E}\left({{\vect{x}_0}}; \phi\right)\right); \theta\right)\right)
\right].  
\end{align}
Given the forward model and the measurements $\bm{y}_{\sf{sem}}$, this problem is  solved  via  imposing a fidelity constraint and a proper prior on the expected ground truth \cite{INN}. Typically, the fidelity constraint is chosen to be MSE, and the prior knowledge can be realized via diffusion generative priors.


\begin{wrapfigure}{r}{0.45\textwidth}
  \centering 
  \vspace{-5mm}
\includegraphics
[width=0.45\textwidth, trim={0.0in 0.0in 0.0in  0.0in},clip]{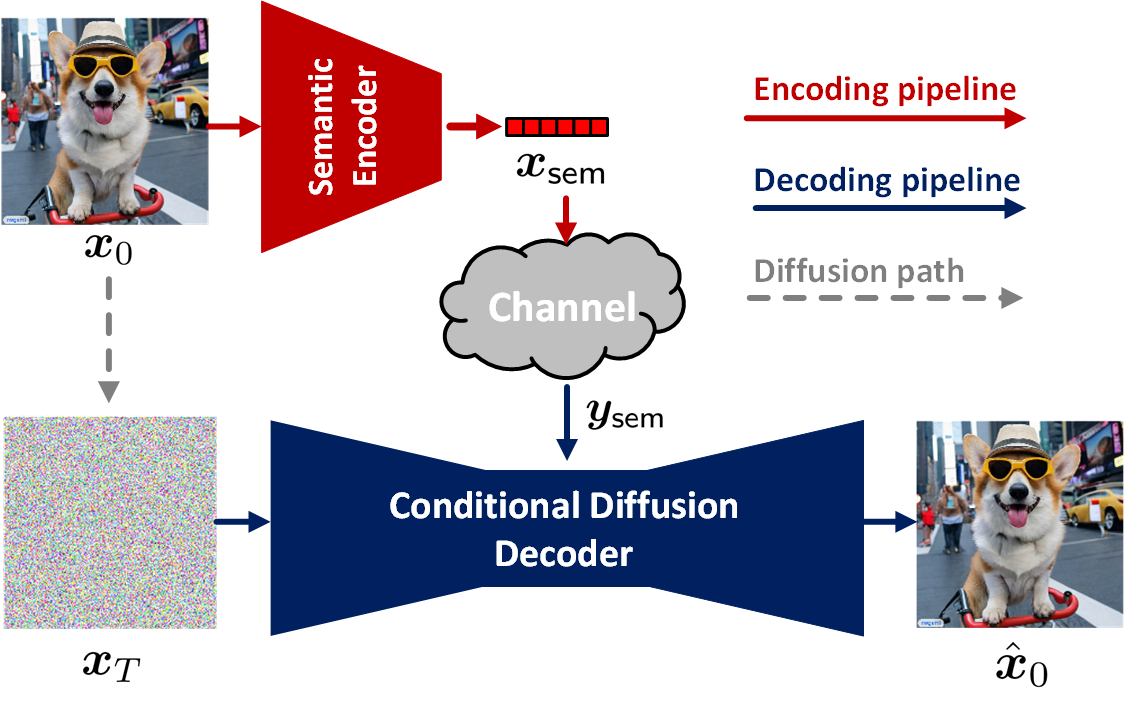} 
  \caption{\small Communication pipeline of the proposed SemCom framework.}
  \label{fig:SysMod}
\end{wrapfigure}  

\section{Solution}  \label{sec:solution}  
Fig. \ref{fig:SysMod} depicts  a schematic of the proposed solution. Semantic encoder extracts the semantics, which are then communicated with the receiver. The semantic latent vector is used for the conditioning of the CDiff decoder, starting the denoising from $\boldsymbol{x}_T$ as input. The intuition behind $\boldsymbol{x}_T$  as input here is that it inherently comes from the noisy (stochastic) version of the ground-truth which the decoder has learned during the forward diffusion training. 
 While the autoencoder architecture  has limitations in  compressing stochastic details, this is smartly handled by the diffusion model that learns to denoise and decode the semantics,  while preserving the fine-grained details.  
 
\subsection{Semantic Encoder Model} \label{subsec:encoder} 
Semantic encoder's functionality is modeled by a mapping  from the source data  to  semantic latent, i.e., $\vect{x}_0 \rightarrow \vect{y}_{\sf sem}$.  
The goal of the encoder  is to summarize high-dimensional source information into a descriptive vector $\vect{y}_\mathsf{sem} = \mathcal{E}(\vect{x}_0; \phi)$  of latent semantic  representations such that it contains the necessary information to assist  the decoder in the denoising and reconstruction process $p_\theta(\vect{x}_{t-1}|\vect{x}_t, \vect{y}_\mathsf{sem})$.   
We do not assume any particular architecture for this encoder; because, thanks to the conditioning mechanism of diffusion models, we do not assume any matched encoder-decoder architecture, and the decoder model can work with varying-length latents. 
In this paper, since we assume that the source data-stream is image-type data, we employed CNN architecture, with details in Appendix \ref{app:neural_architecture_encoder}.

\subsection{Diffusion-Based Semantic Decoder Model}
\label{sec:deocder}   
The main focus of the proposed scheme is on the design of the decoder model. 
The CDiff, realized via a conditional DDPM,  first goes through a forward diffusion process, i.e., the Markov chain of $(\vect{x}_0; \ysem$) --- $(\vect{x}_t; \ysem)$ ---  $\vect{x}_T$,  purposefully diffusing the ground-truth samples by adding noise. 
That is, the decoder first aims to learn the finer-grained details of source  information, which are not expressed by $\vect{y}_{\sf sem}$  due to the limited capacity of the neural encoder.  
Then in a reverse process, the CDiff decoder  
 de-maps back to the  ground-truth probability distribution space, starting from $\vect{x}_T$ and guided by the semantics $\vect{y}_{\sf sem}$. I.e.,    
$(\vect{y}_{\sf sem}; \vect{x}_T)$ --- $(\vect{y}_{\sf sem}; \vect{x}_t)$ ---  $\vect{x}_0$.      
The decoding transition probability 
can be modeled  by a conditional  probability distribution  $p_\theta(\xtone|\xt, \ysem)$.   The proposed CDiff decoder  
is expected to estimate the reverse transition probability distribution that match the true distribution $q(\vect{x}_{t-1}|\vect{x}_t, \vect{x})$.  
The  reverse process of ``generative decoding'' can then be modeled as   
\begin{align} 
    p_\theta(\vect{x}_{t-1} | \xt, \ysem) & \approx q(\vect{x}_{t-1}|\vect{x}_t, \vect{x}_0), \label{eq:rev} 
   \overset{(a)}{\approx}  q\left(\vect{x}_{t-1}|\vect{x}_t, \hat{{\vect x}}_\theta(\vect x_t, \ysem; t)\right),   \\ 
   p_\theta(\boldsymbol{x},\boldsymbol{x}_{1:T}| \ysem) &= p(\vect{x}_T) \prod_{t=1}^Tp_\theta(\vect{x}_{t-1} | \vect{x}_t, \vect{y}_\text{sem}), 
\end{align}
where  
the true distribution $q(\vect{x}_{t-1}|\vect{x}_t, \vect{x})$ is approximated in $(a)$ by replacing the unknown ground-truth ${\vect x}_0$ with its corresponding estimate $\hat{{\vect x}}_\theta(\vect x_t, \ysem; t)$.

\paragraph{Forward  process \& training}
Having the semantic information to guide the decoder via conditioning,  the  forward  diffusion process can be modeled via the following transition probability   
\begin{align}  
  q({\vect x}_t|{{\vect x}}_0,{\widetilde{\vect y}_{\sf sem}}) \sim  \mathcal{N}\Big((1-w_t)\sqrt{\bar{\alpha}_t}{{\vect x}}_0 + w_t\sqrt{\bar{\alpha}_t}\widetilde{\vect y}_{\sf sem},\delta_t{\bm I}\Big),
  \label{eq:new diffusion process t}
\end{align} 
where  $\delta_t := (1-\bar{\alpha}_t)-w_t^2\bar{\alpha}_t$, and $\widetilde{\vect y}_{\sf sem}$  is the padded-and-reshaped version of the conditional information, to have the same dimensions as the source samples, to enable distilling the semantics into the learning process.   This is a generalization of the vanilla  forward diffusion process, where a weighted sum (parameterized by $w_t$) of the ground-truth and the conditioning is considered. 
Typically, $w_t$ start from $0$ and  gradually increases to $w_T \approx 1$ via a pre-defined scheduling. Intuitively, this guides the model during each step of  training, to first focus  on diffused ground-truth samples, explore, and learn their fine-grained details; and then, while the diffused samples loose their expressiveness due to high volume of noise, the model switches its attention to learning from the semantic latents.

 Training is then carried out by optimizing the following loss function. 
\begin{equation}
\mathcal{L}({{\theta}, \phi}) = 
\mathbb{E}_{\subalign{&t \sim \mathsf{Unif}[T] \\ &{\vect x}_0, \ysem \sim p(\vect{x}_0, \ysem) \\ &{\vect x}_t \sim q(\vect{x}_t | \vect{x}_0, \ysem)}}  \Big[ 
\norm{\hat{{\vect x}}_\theta({{\vect x}}_t, \ysem; t) - {\vect{x}_0}}_2^2
\Big]. 
\label{eq:loss}
\end{equation}

\paragraph{Sampling process} 
Each step  of the decoding (so-called 
``sampling'')  is viewed  as an interpolation between the denoised samples ${\vect x}_t$, the conditional information regarding the source data (semantic latents in our case)    $\ysem$, and the predicted noise $\bm \epsilon_{\theta}(\cdot, t)$: 
\begin{equation}
   {\vect x}_{t-1} = \psi_{x}{\vect x}_{t} + \psi_{y} \ysem - \psi_{\epsilon} {\bm \epsilon}_\theta({\vect x}_t, \ysem, t) + \sqrt{\delta_t} {\vect z}, \quad {\vect z} \sim \mathcal{N}(\bm 0, I)
   \label{eq:new reverse mean}
\end{equation} 
where ${\bm \epsilon}_\theta({\vect x}_t, \ysem, t)$ is the noise predictor model  obtained by  reparameterization of $\hat{{\vect x}}_\theta({{\vect x}}_t, \ysem; t)$:   
\begin{align}
{\vect \epsilon}_\theta({\vect x}_t, \ysem, t)  := \frac{\left(\vect{x}_t - \sqrt{\bar{\alpha}_t} \hat{{\vect x}}_\theta({{\vect x}}_t, \ysem; t)\right)}{\sqrt{1-\bar{\alpha}_t}}.
\end{align}  
 The coefficients $ \psi_{x}, \psi_y, $ and $ \psi_{\epsilon}$ can be estimated  by solving the evidence lower bound (ELBO) optimization criterion on the objective function following \cite{DM_Ho}. In this work, we use the coefficients as derived in \cite{cdiff}, where   
$
    \psi_{x} = \frac{\delta_{t-1} (1-\lambda_t)}{\delta_{t} (1-\lambda_{t-1})}\sqrt{\alpha}_t + (1-\lambda_{t-1})\frac{\delta_{t|t-1}}{\delta_{t}\sqrt{\alpha}_t}$,
   $ \psi_{y} = (\lambda_{t-1}\delta_t - \frac{\lambda_t(1-\lambda_t)}{1-\lambda_{t-1}}\alpha_t\delta_{t-1})\frac{\sqrt{\bar{\alpha}_{t-1}}}{\delta_t}$, 
$    \psi_{\epsilon} = (1-\lambda_{t-1})\frac{\delta_{t|t-1}\sqrt{1-\bar{\alpha}_t}}{\delta_t \sqrt{\alpha_t}}$, and 
 $\delta_{t|t-1} {=} \delta_t - \left(\frac{1-\lambda_t}{1-\lambda_{t-1}}\right)^2 \alpha_t \delta_{t-1}$.

 \subsection{Theoretical Analysis} 
\label{sec:consistency}

\begin{theorem}
    Assume $\theta^\ast_n$ be a minimizer of an $n$-sample Monte
Carlo approximation of $\mathcal{L}({\theta}, \phi)$ with respect to the decoder model $\theta$. The conditional diffusion decoder $\hat{{\vect x}}_{\theta^\ast_n}({{\vect x}}_t, \ysem; t)$ is a consistent estimator of the ground-truth, i.e., for sufficiently large number of Monte Carlo samples, we have 
    \begin{align}
       \hat{{\vect x}}_{\theta^\ast_n}({{\vect x}}_t, \ysem; t) \overset{{P}} {\longrightarrow} \vect x_0.   
    \end{align} 
\end{theorem} 
\vspace{-5mm}
\begin{proof} 
Please see Appendix \ref{app:proof} 
\end{proof}



\section{Experiments}     \label{sec:results}
\subsection{Results over MNIST}  

\begin{figure*}[tbph]
\centering
    \begin{subfigure}{0.24\textwidth}
        \centering    
        \includegraphics[width=\linewidth, trim={0.0in 0.0in 0.0in  0.0in},clip]{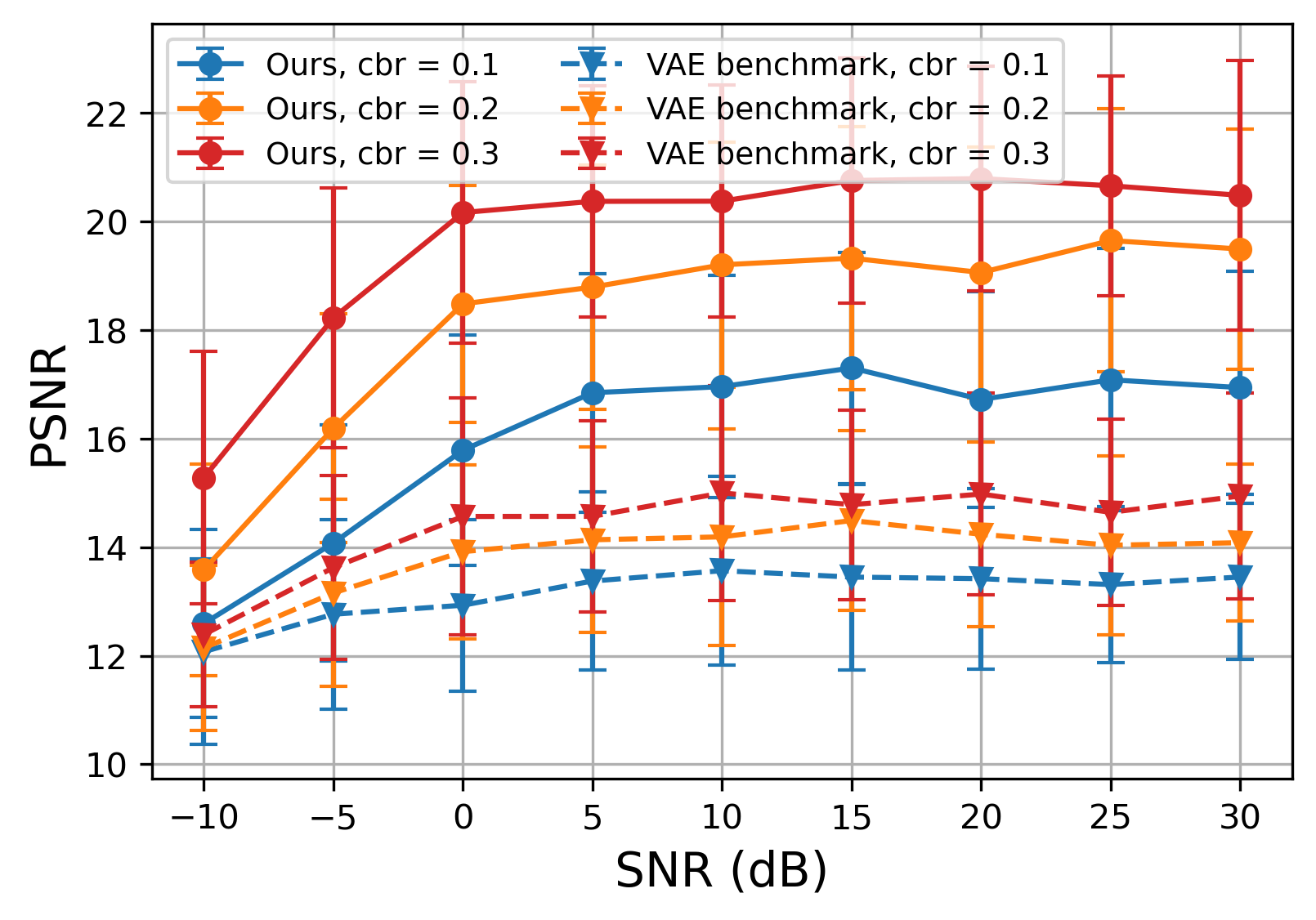}
        \caption{\small}
    \end{subfigure}
    \begin{subfigure}{0.24\textwidth}
        \centering    
        \includegraphics[width=\linewidth, trim={0.0in 0.0in 0.0in  0.0in},clip]{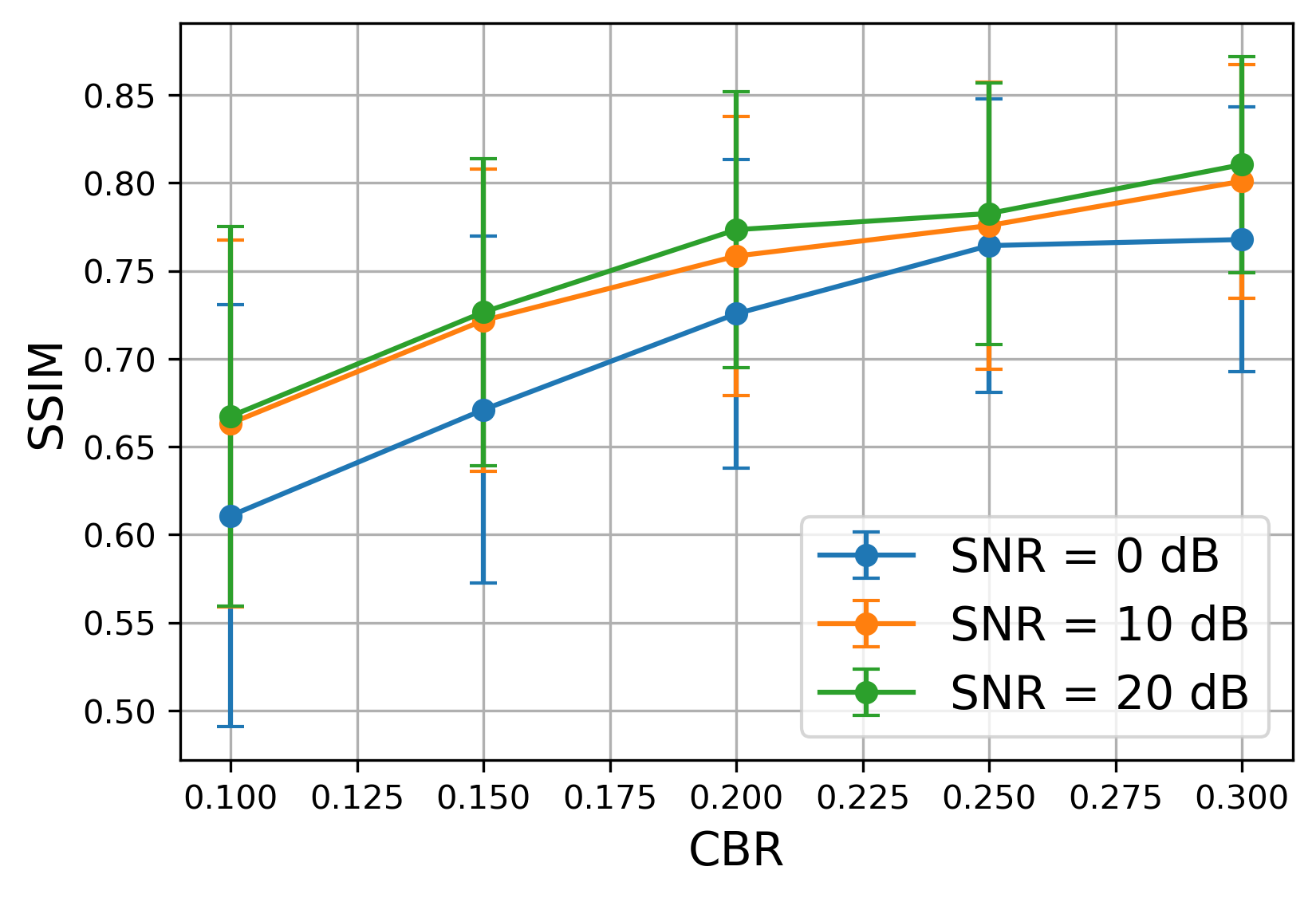}
        \caption{\small}
    \end{subfigure}
    \begin{subfigure}{0.24\textwidth}
        \centering    
        \includegraphics[width=\linewidth, trim={0.0in 0.0in 0.0in  0.0in},clip]{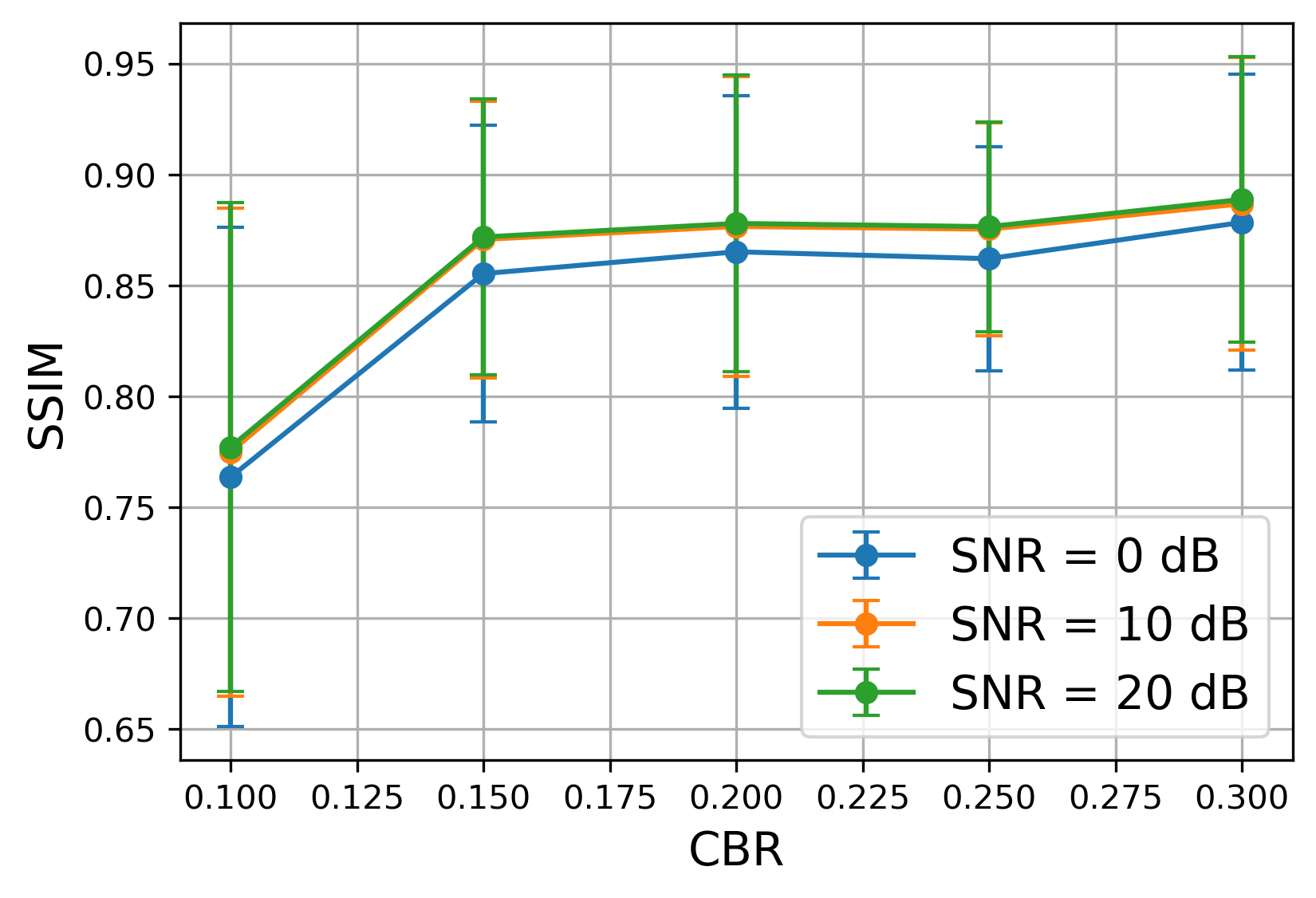}
        \caption{\small}
    \end{subfigure}
    \begin{subfigure}{0.24\textwidth}
        \centering    
        \includegraphics[width=\linewidth, trim={0.0in 0.0in 0.0in  0.0in},clip]{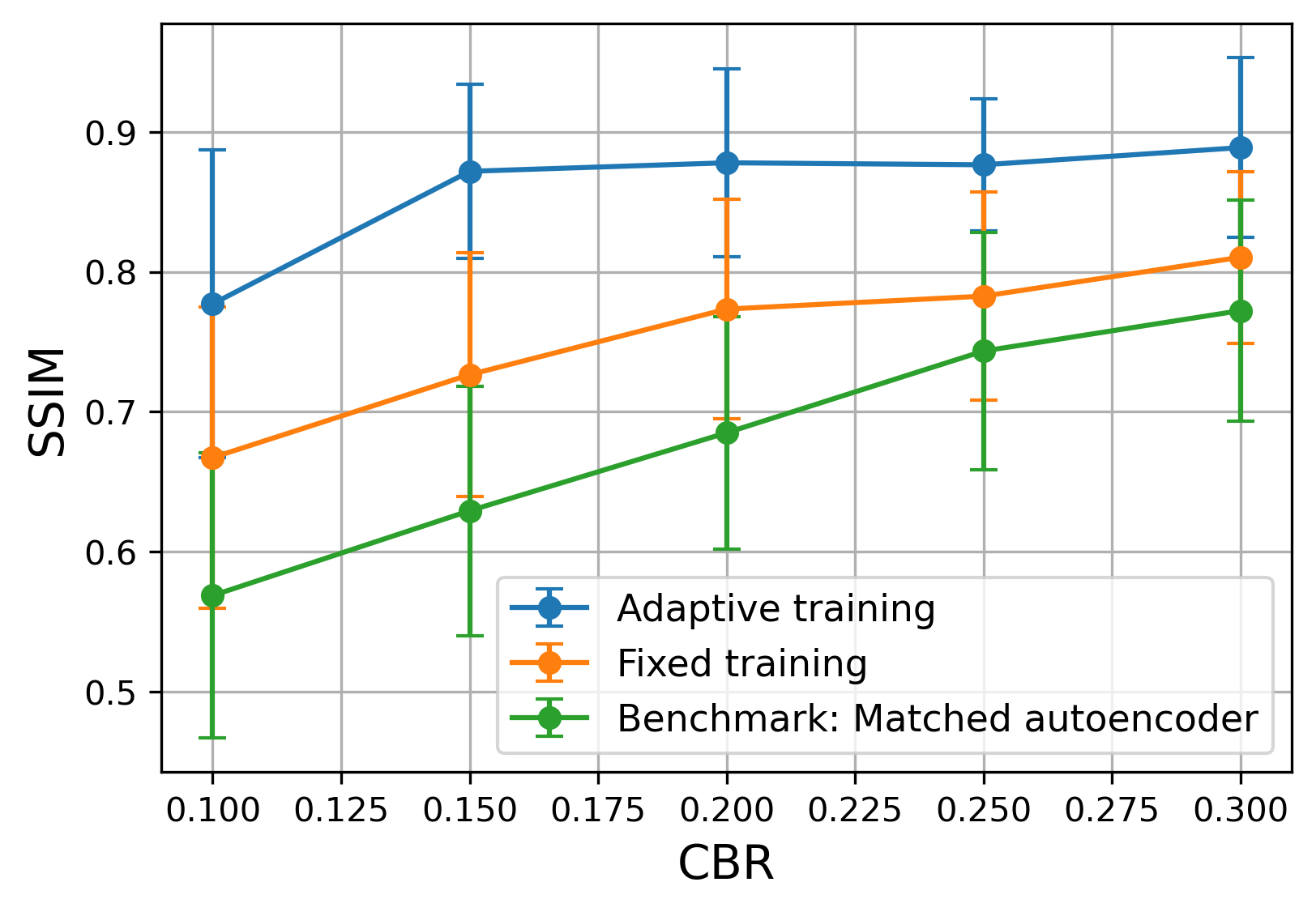}
        \caption{\small}
    \end{subfigure}
    \caption{\small Effects of SNR and CBR on SemCom performance for different training strategies over MNIST.  
    }
    \label{fig:mnist_snr_cbr} 
    \vspace{0mm}
\end{figure*}

\begin{figure*}[tbph]
\centering
    \begin{subfigure}{0.24\textwidth}
        \centering    
        \includegraphics[width=\linewidth, trim={0.0in 0.0in 0.0in  0.0in},clip]{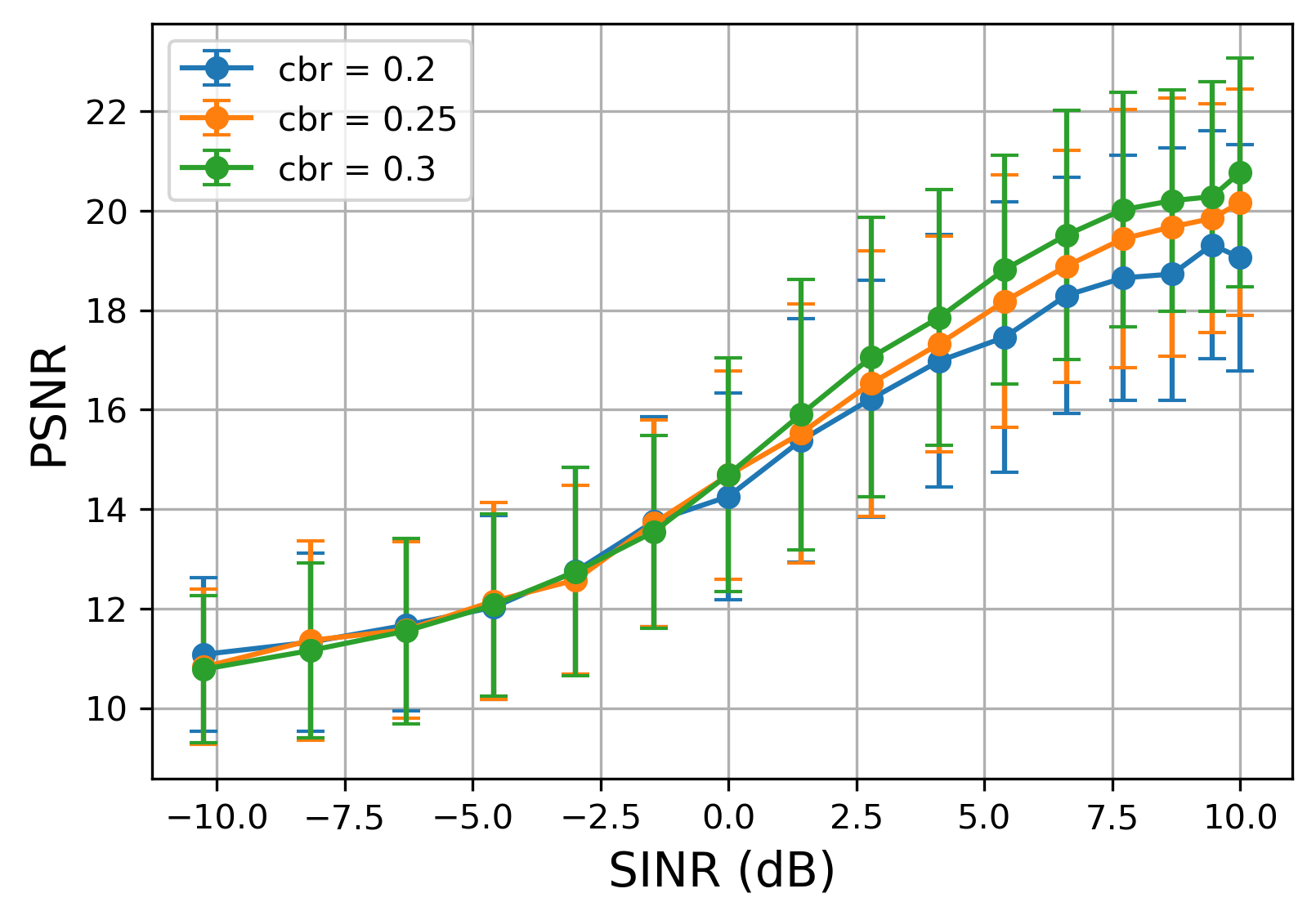}
        \caption{\small}
    \end{subfigure}
    \begin{subfigure}{0.24\textwidth}
        \centering    
        \includegraphics[width=\linewidth, trim={0.0in 0.0in 0.0in  0.0in},clip]{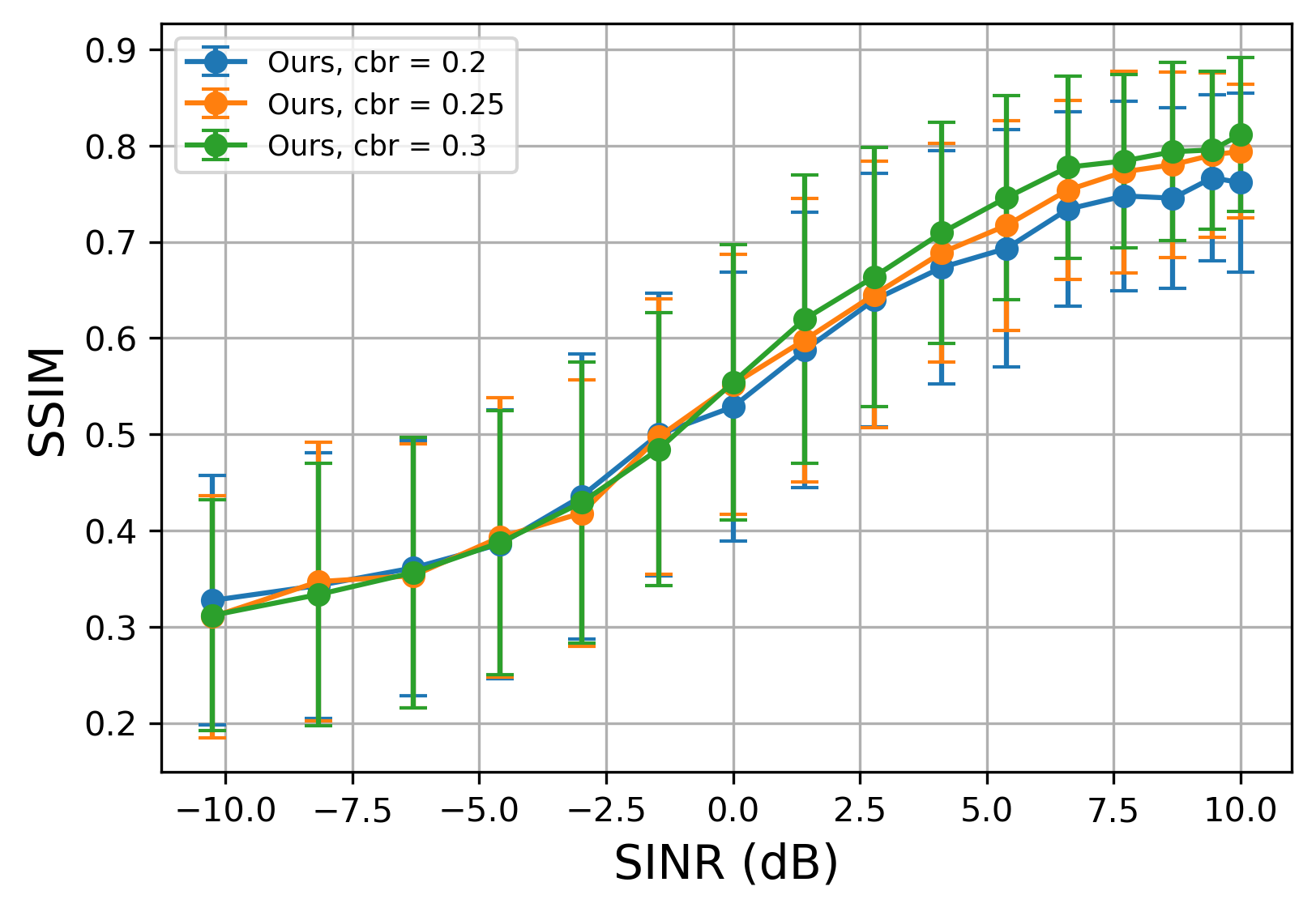}
        \caption{\small}
    \end{subfigure}
    \begin{subfigure}{0.24\textwidth}
        \centering    
        \includegraphics[width=\linewidth, trim={0.0in 0.0in 0.0in  0.0in},clip]{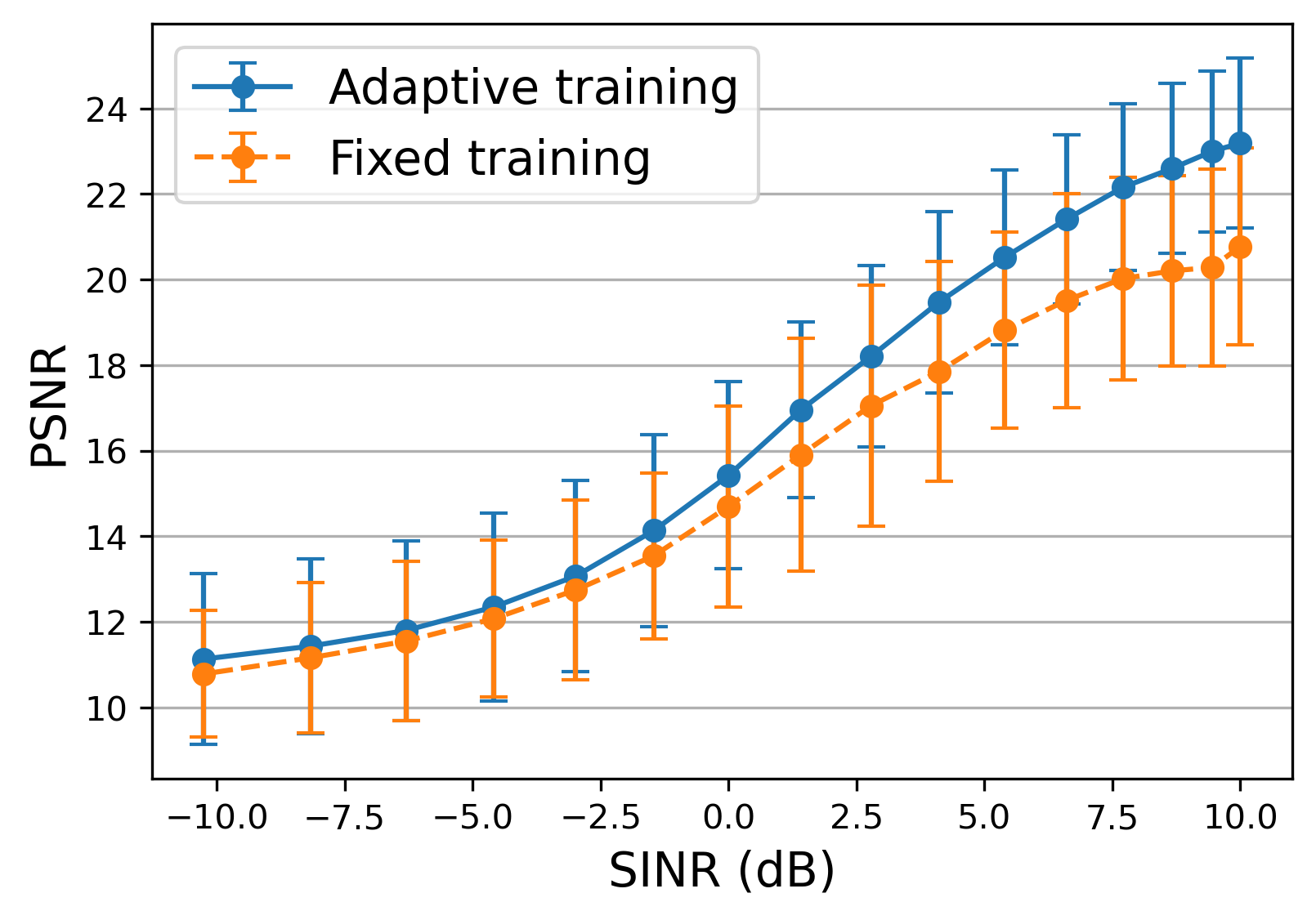}
        \caption{\small}
    \end{subfigure}
    \begin{subfigure}{0.24\textwidth}
        \centering    
        \includegraphics[width=\linewidth, trim={0.0in 0.0in 0.0in  0.0in},clip]{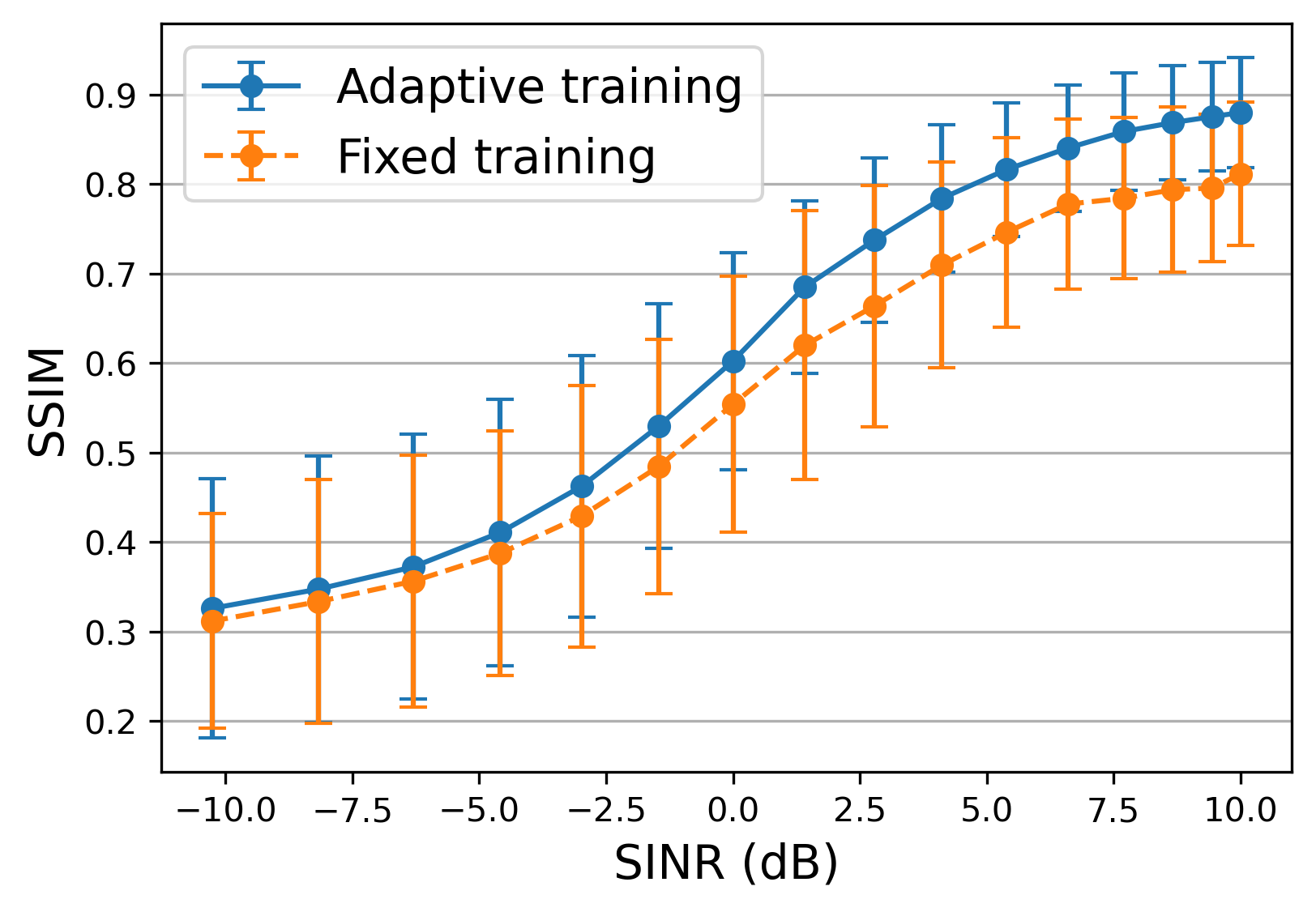}
    \caption{\small}
    \end{subfigure}
    \caption{\small Effects of multi-user interference (in terms of SINR) on SemCom performance over MNIST.  
    }
    \label{fig:mnist_interf}  
    \vspace{0mm}
\end{figure*}

\vspace{-2mm}
Fig. \ref{fig:mnist_snr_cbr} illustrates the  reconstruction performance of our scheme over a wide range of channel SNRs and CBR values. In Fig. \ref{fig:mnist_snr_cbr}-(a) we start with studying the very common reconstruction performance in terms of the PSNR. 
Notably, our scheme reasonably shows the capability to improve its performance when the CBR or SNR is increased, while the VAE benchmark does not show any notable performance in response to the improvement in channel bandwidth or SNR.    We further remark that the scheme has been trained over SNR values between $-10$ dB to $10$ dB, and a fixed CBR of 0.3 (for fixed-training strategy) while showcasing promising results over  a wide range of channel SNRs and  CBR values. 
Please see Appendix \ref{app:train} for the details on training and hyperparameters.    
In Figs. \ref{fig:mnist_snr_cbr}-(b) to \ref{fig:mnist_snr_cbr}-(d),  the effect of CBR is studied.    
Fig. \ref{fig:mnist_snr_cbr}-(b) evaluates the performance of the model when trained for a fixed CBR (0.3), while Fig. \ref{fig:mnist_snr_cbr}-(c) evaluates the performance when the model is trained over a set of defined CBRs (see Appendix \ref{app:train} for the details). Fig. \ref{fig:mnist_snr_cbr}-(d) compares the results with the matched autoencoder architecture, highlighting outperforming  of our scheme especially for radio resource limited scenarios (i.e., low CBRs, thus low available bandwidth).    
Comparing Figs. \ref{fig:mnist_snr_cbr}-(b) and \ref{fig:mnist_snr_cbr}-(c), it can be observed that training the model over a range of CBRs can make the model more robust  (less variations) against changes in channel SNR and CBR. 
These curves provide important design guidelines: One can choose among different available CBRs, depending on how much physical resources are available, and how accurate the reconstruction quality is aimed to be.   
For example, only $10\%$ SSIM improvement can be achieved  if we have three times more communication resources in terms of channel bandwidth  (CBR of 0.1 versus 0.3). One can trade-off whether the SemCom performance is important, or it is preferred to save radio resources (channel bandwidth) by $1/3$.    

In Fig. \ref{fig:mnist_interf}, we extend the simulations for the scenario of multi-user SemCom. This is simulated by having our CDiff decoder receive a convex combination of semantic latents from two neural encoders each independently transmitting their own i.i.d  data samples  over the  same channel.   The figure  studies the effect of multi-user interference measured in terms of the signal-to-interference-plus-noise ratio (SINR) on the reconstruction  performance in terms of PSNR and SSIM.   
While CBR seems to be the most important factor in most of deepJSCC/SemCom papers,  this is the case when the typical  point-to-point single user scenario is considered. However, in a practical system, there might be multiple encoders for multiple intended users, whose latent vectors (semantic meanings) might interfere.  Fig. \ref{fig:mnist_interf} indicates that the interference signals seem to be the dominating factor in a practical multi-user setup, not the CBR. This is because the PSNR/SSIM metrics vary with SINR, (not with the change in CBR).  
This can be a very important practical insight for SemCom system designs.   

 \subsection{Results over CIFAR-10}
 
\begin{figure*}[tbph]
\centering
    \begin{subfigure}{0.24\textwidth}
        \centering    
        \includegraphics[width=\linewidth, trim={0.0in 0.0in 0.0in  0.0in},clip]{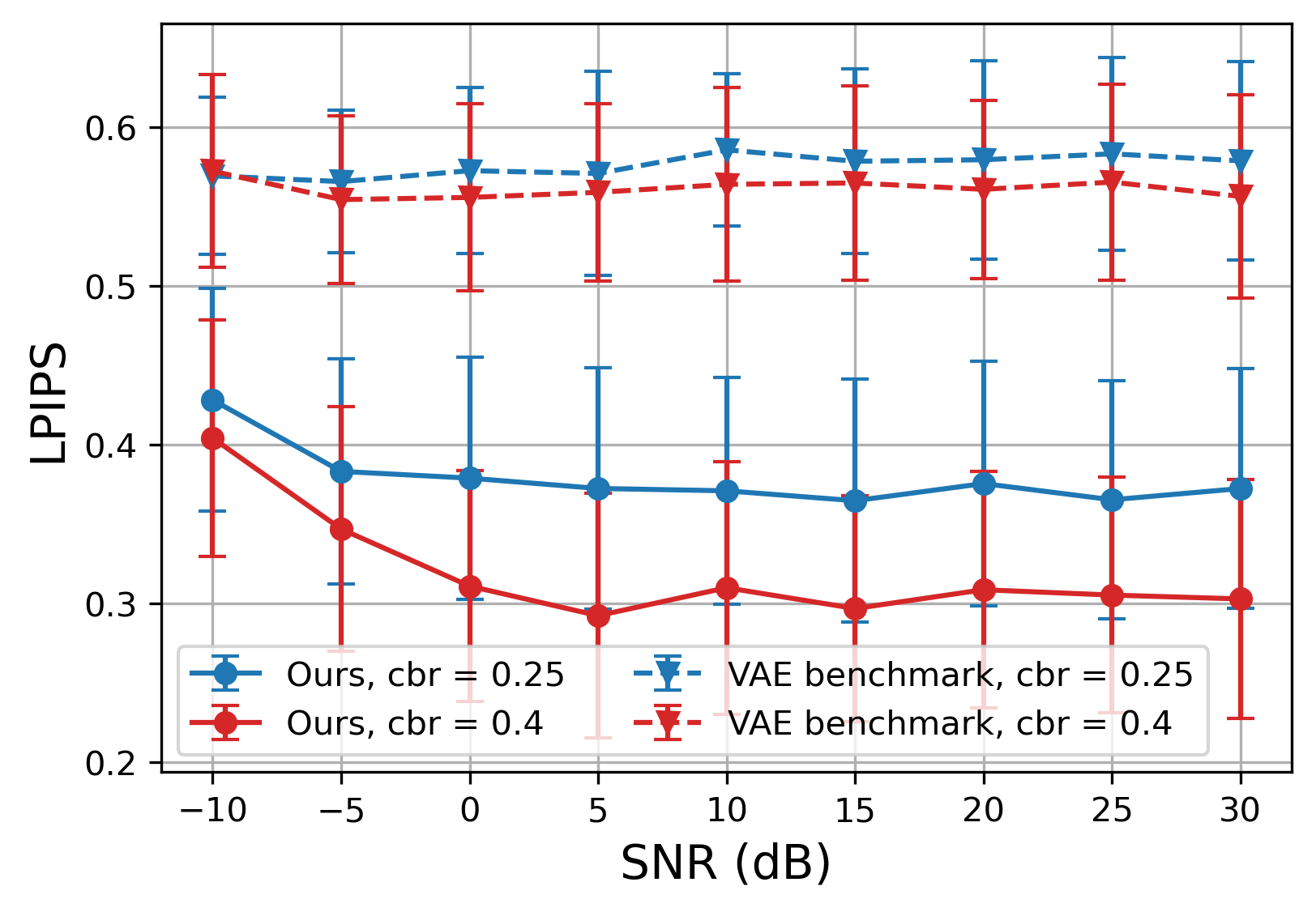}
        \caption{\small}
    \end{subfigure}
    \begin{subfigure}{0.24\textwidth}
        \centering    
        \includegraphics[width=\linewidth, trim={0.0in 0.0in 0.0in  0.0in},clip]{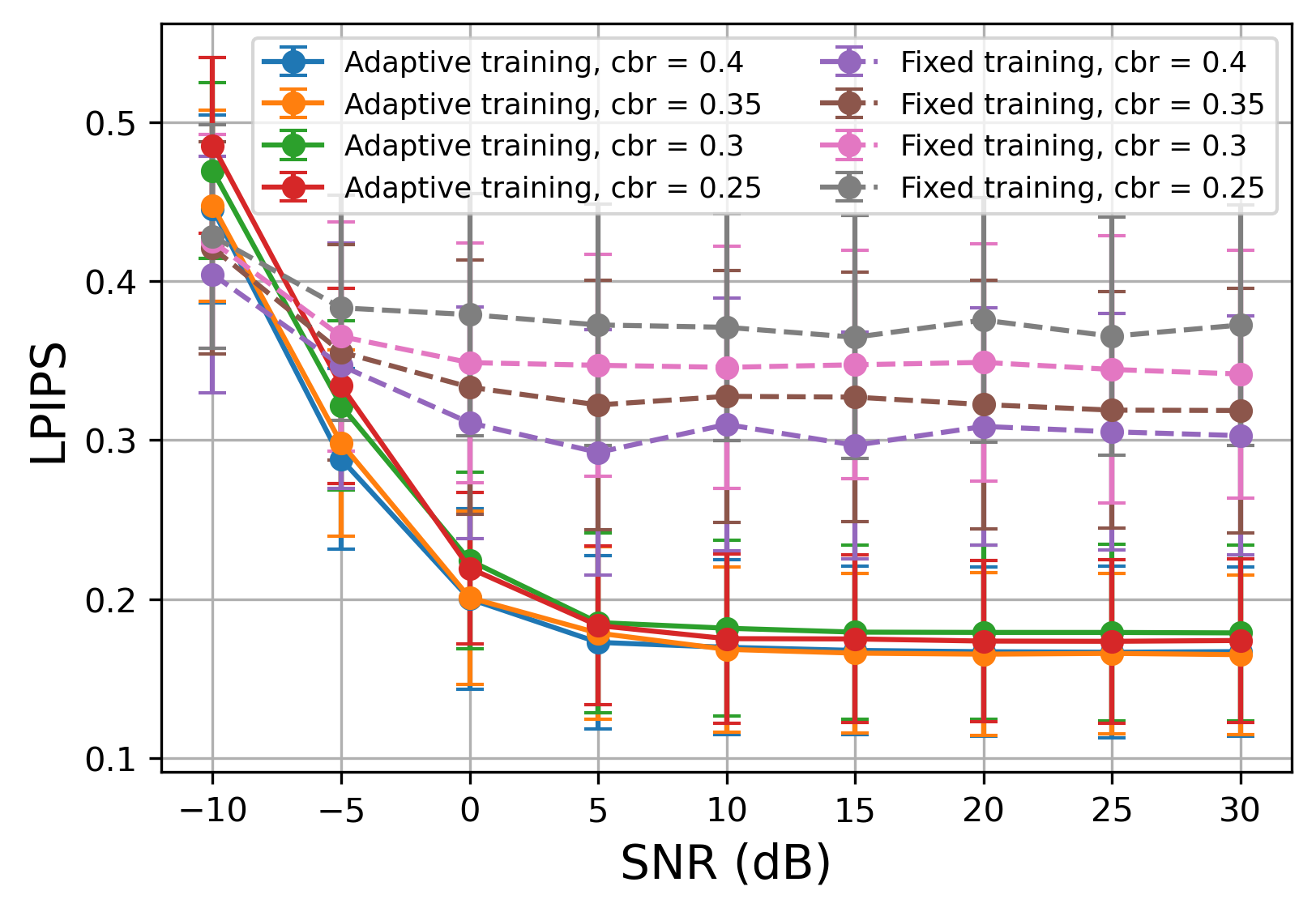}
        \caption{\small}
    \end{subfigure}
     \begin{subfigure}{0.24\textwidth}
        \centering    
        \includegraphics[width=\linewidth, trim={0.0in 0.0in 0.0in  0.0in},clip]{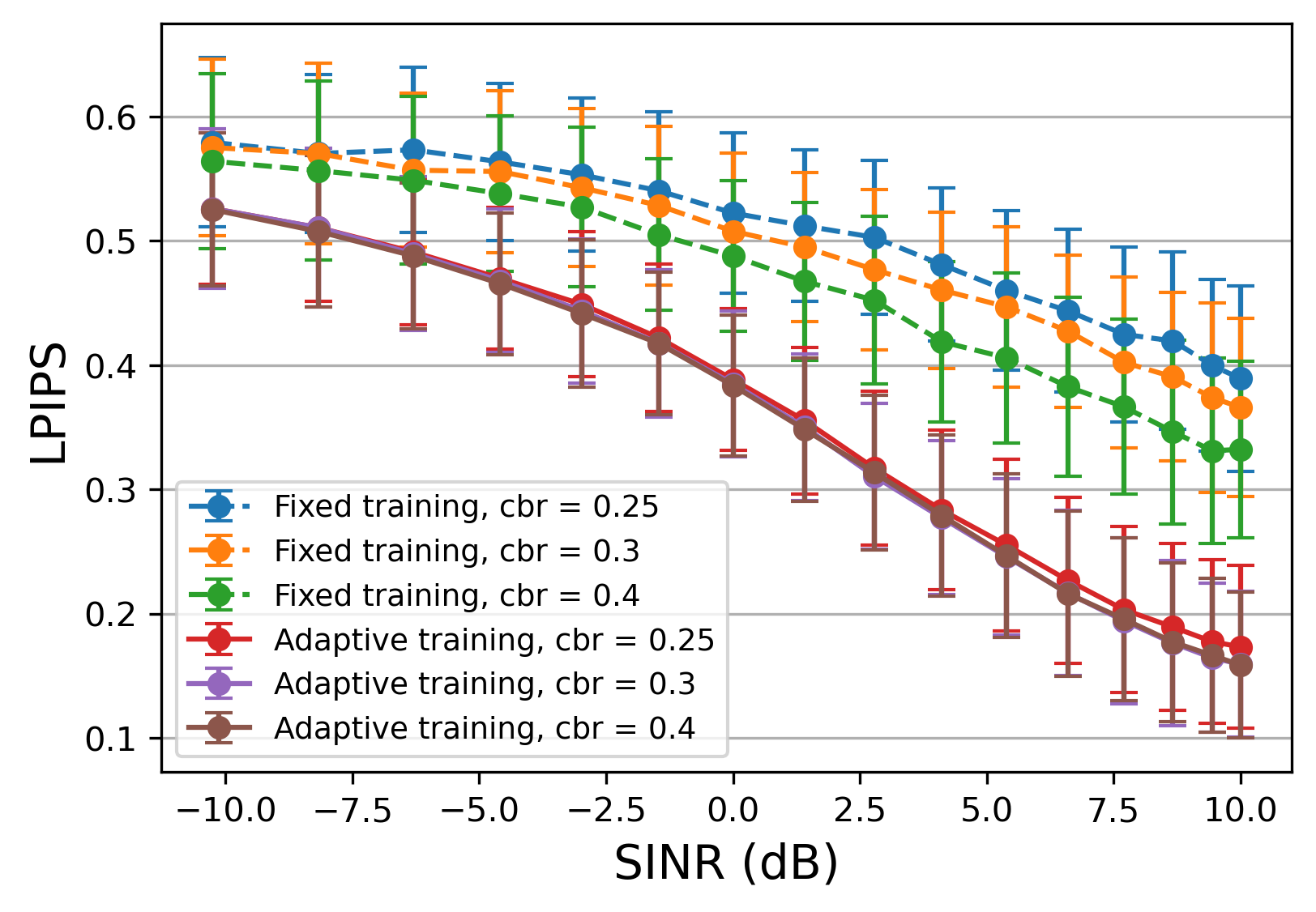}
        \caption{\small}
    \end{subfigure}
    \begin{subfigure}{0.24\textwidth}
        \centering    
        \includegraphics[width=\linewidth, trim={0.0in 0.0in 0.0in  0.0in},clip]{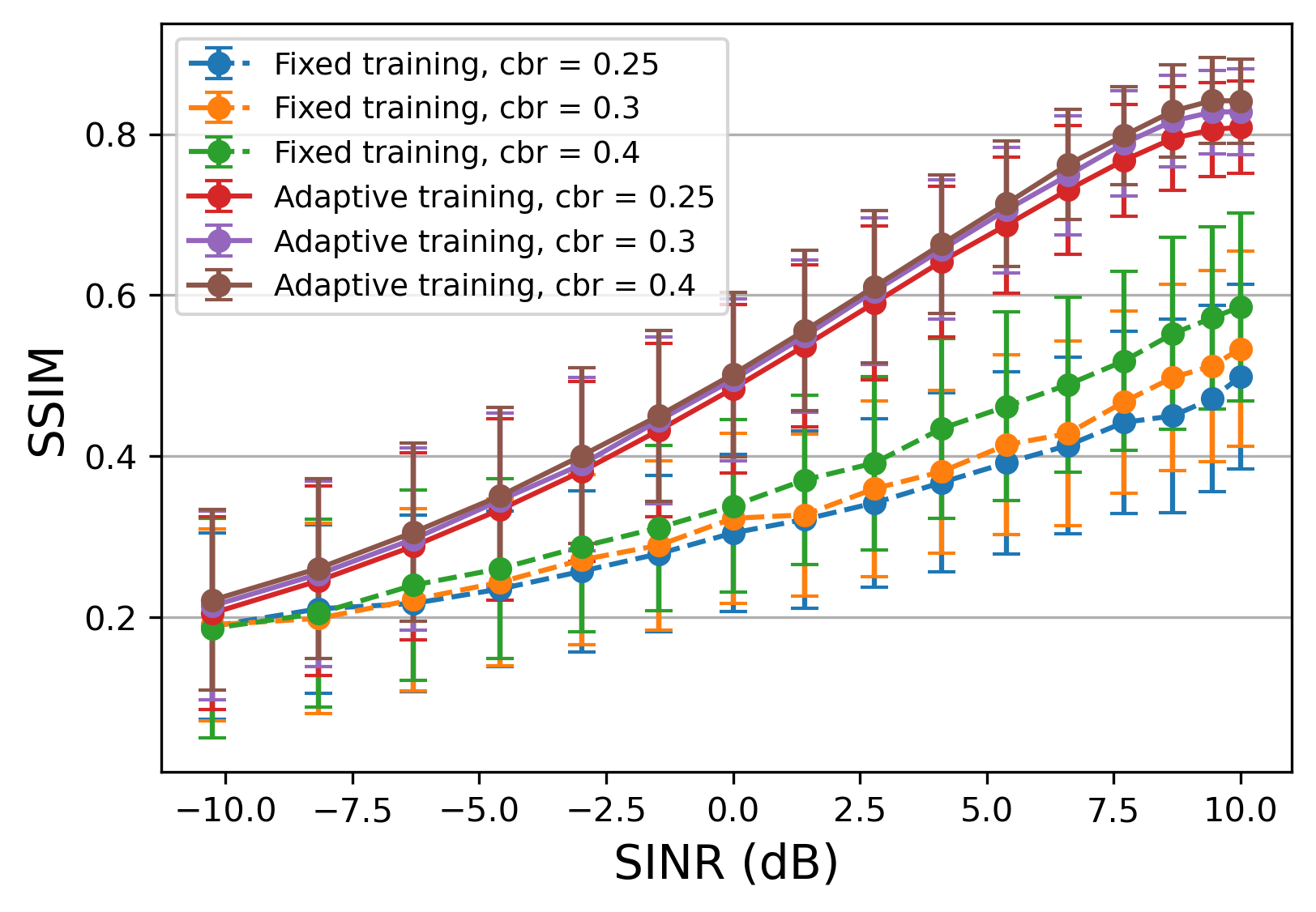}
        \caption{\small}
    \end{subfigure}
    \caption{\small Effects of SNR and SINR on SemCom performance for different training strategies over CIFAR-10.  
    }
    \label{fig:cifar_snr_sinr} 
    \vspace{0mm}
\end{figure*}

 \begin{figure*}[tbph]
\centering
    \begin{subfigure}{0.24\textwidth}
        \centering    
        \includegraphics[width=\linewidth, trim={0.0in 0.0in 0.0in  0.0in},clip]{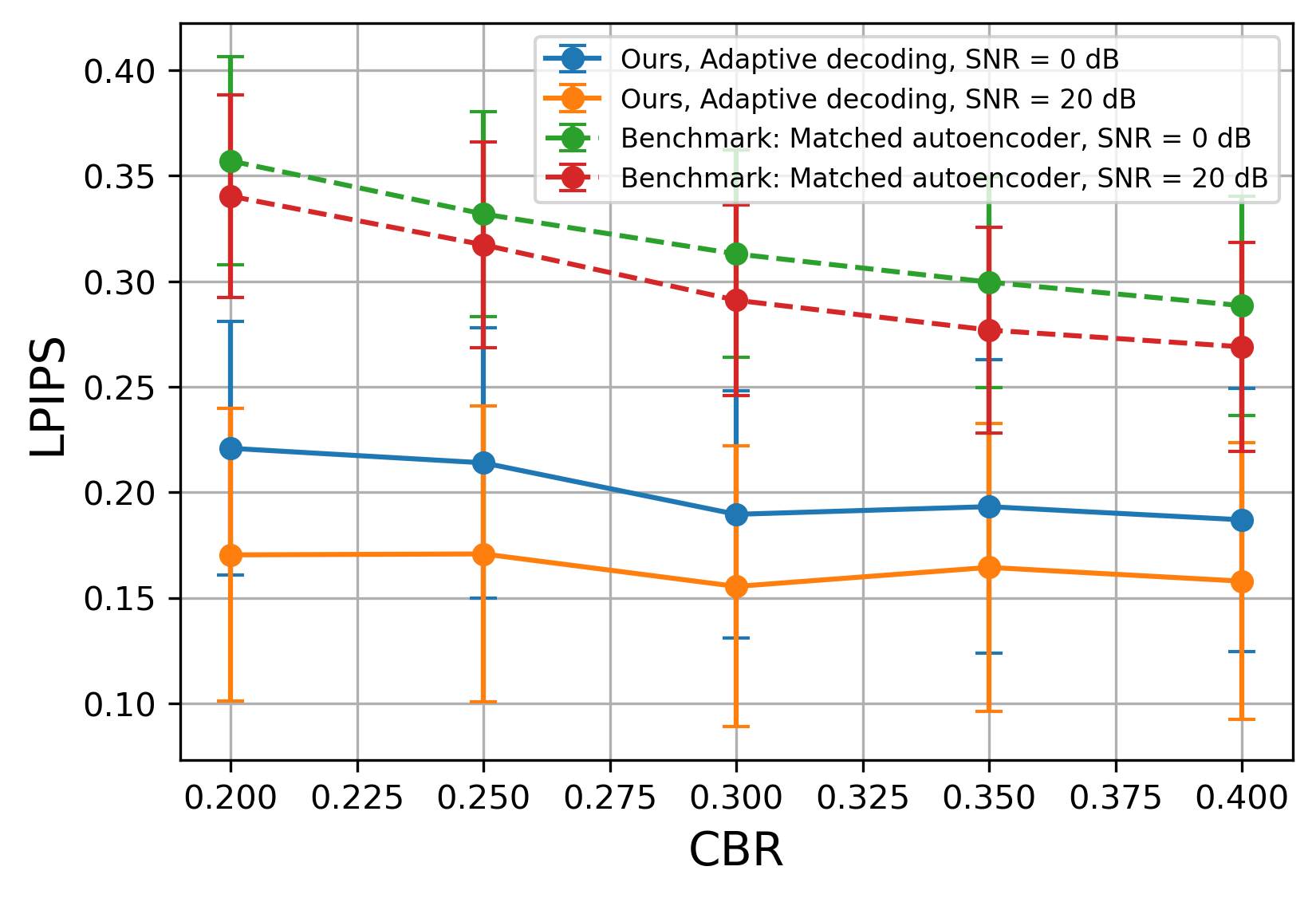}
        \caption{\small}
    \end{subfigure}
    \begin{subfigure}{0.24\textwidth}
        \centering    
        \includegraphics[width=\linewidth, trim={0.0in 0.0in 0.0in  0.0in},clip]{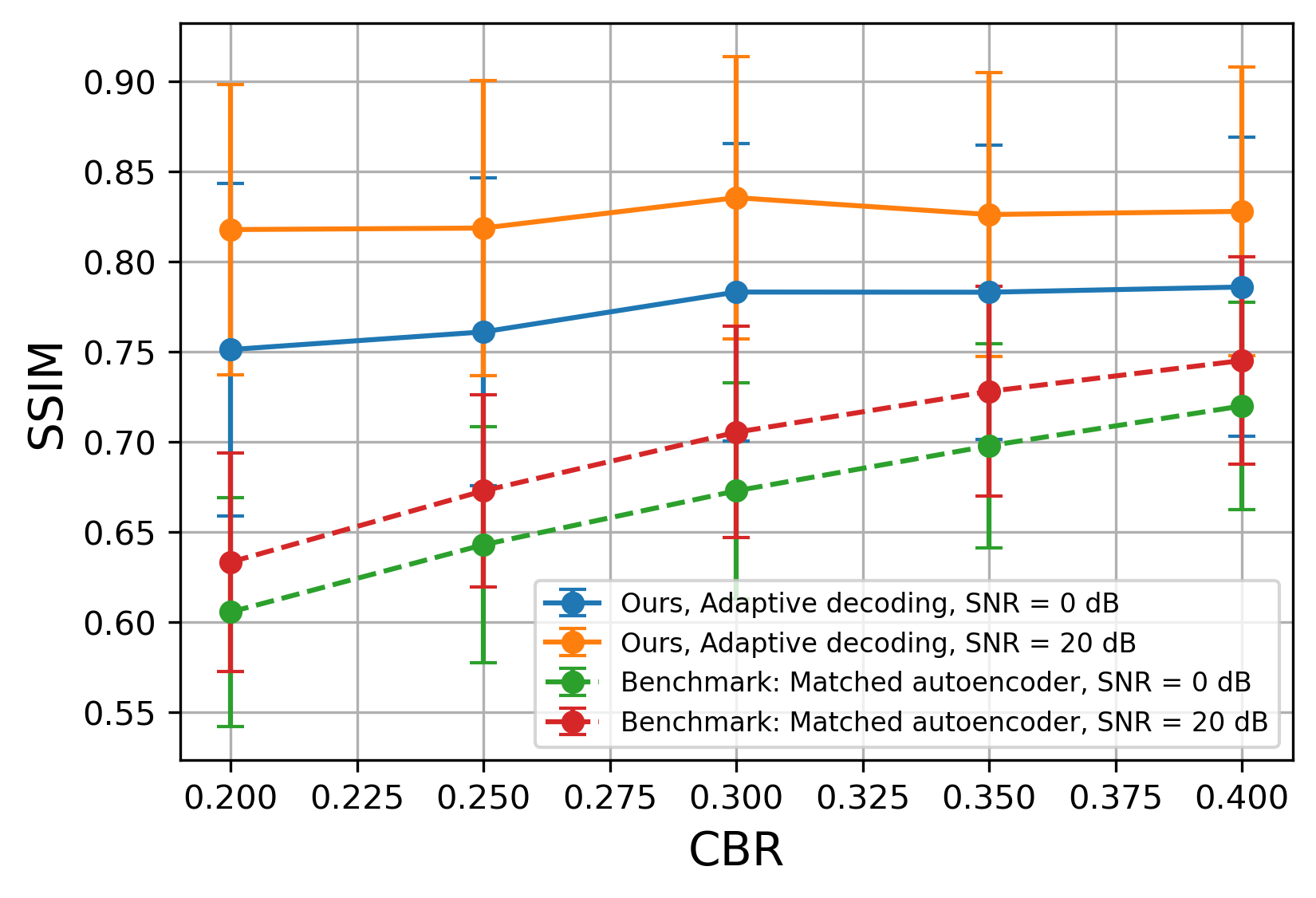}
        \caption{\small}
    \end{subfigure}
    \begin{subfigure}{0.24\textwidth}
        \centering    
        \includegraphics[width=\linewidth, trim={0.0in 0.0in 0.0in  0.0in},clip]{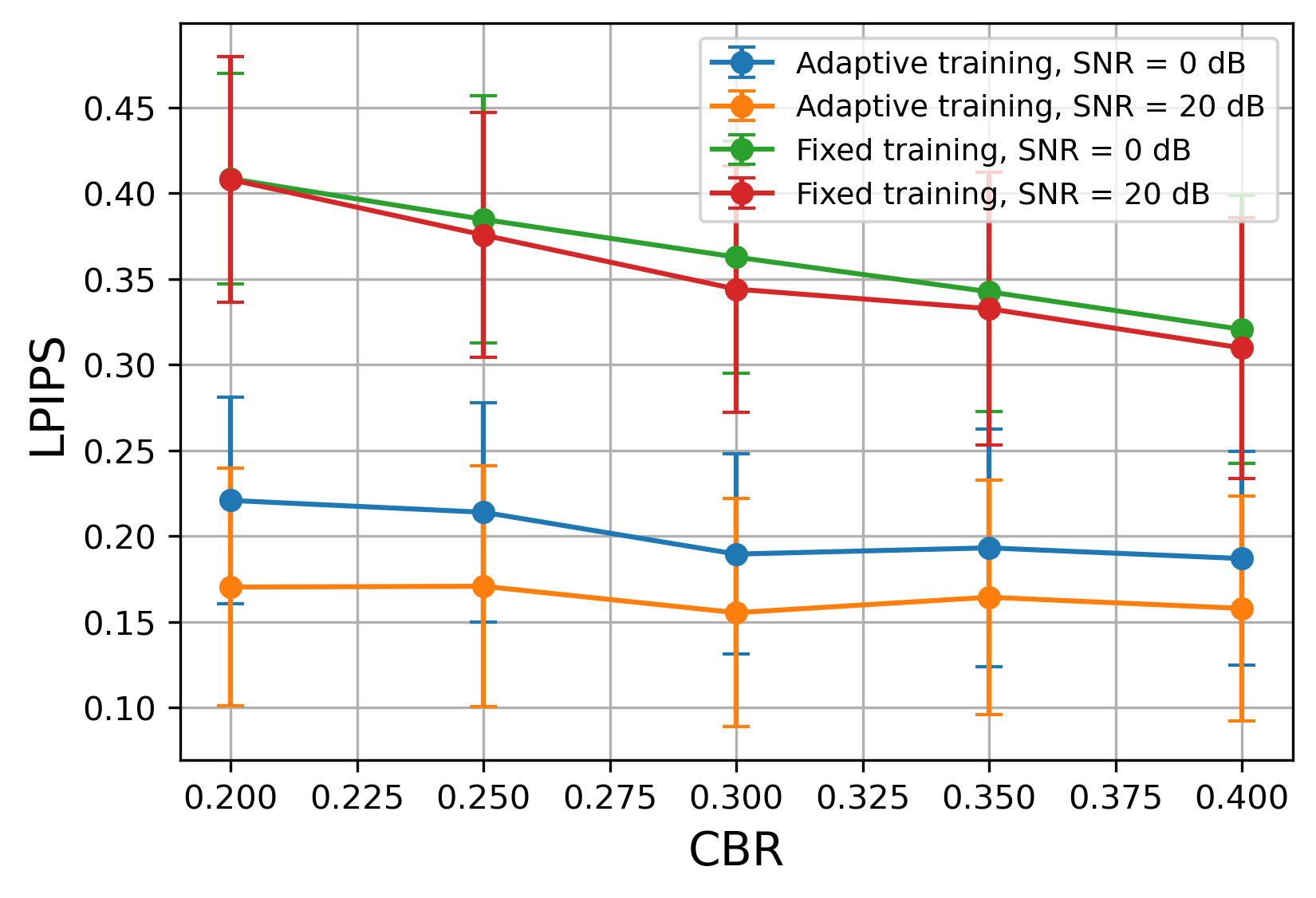}
        \caption{\small}
    \end{subfigure}
    \begin{subfigure}{0.24\textwidth}
        \centering    
        \includegraphics[width=\linewidth, trim={0.0in 0.0in 0.0in  0.0in},clip]{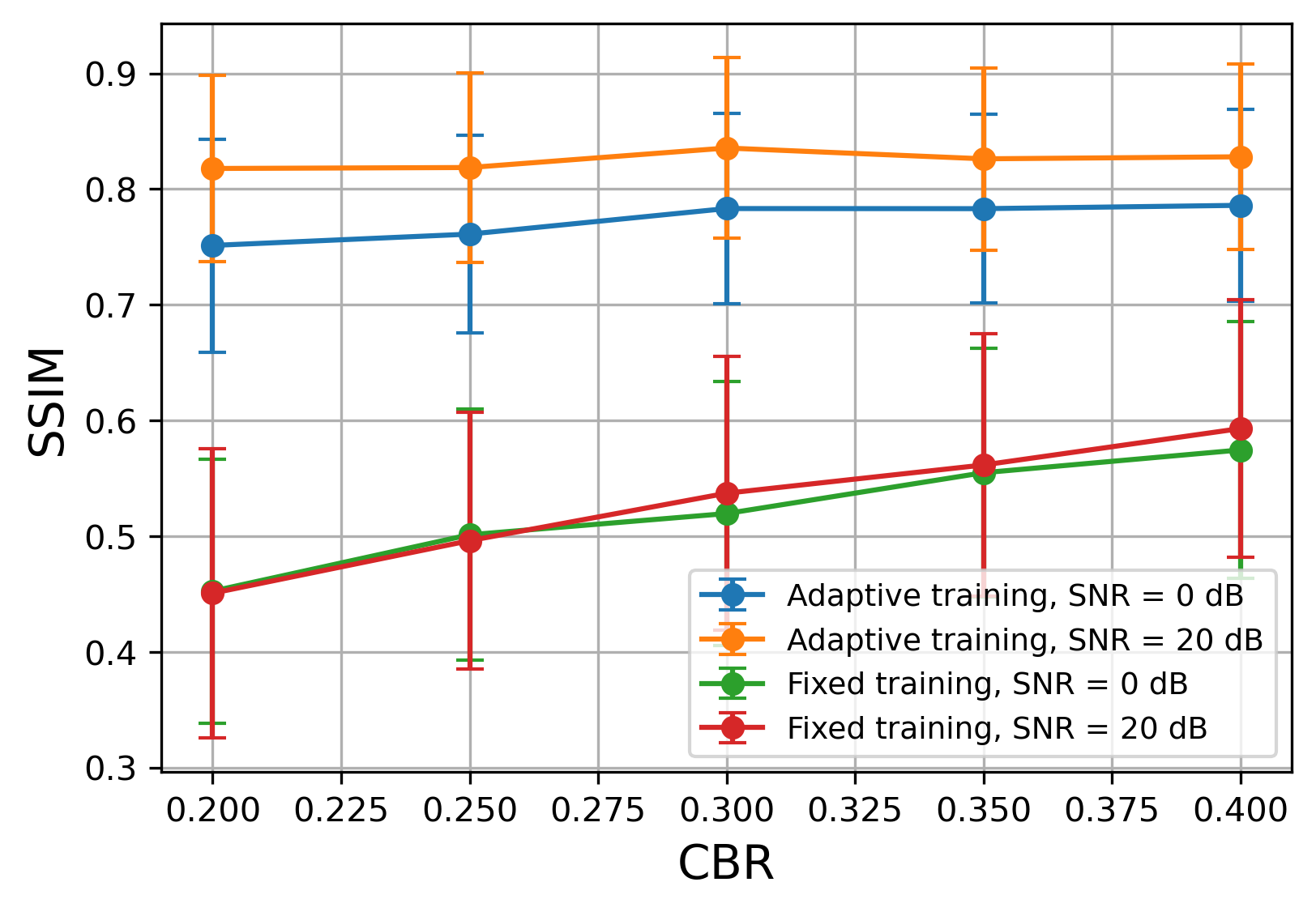}
        \caption{\small}
    \end{subfigure}
    \caption{\small Effects of CBR on SemCom performance over CIFAR-10.  
    }
    \label{fig:cifar_cbr} 
    \vspace{0mm}
\end{figure*}

Fig. \ref{fig:cifar_snr_sinr} studies the  performance  with respect to  LPIPS and SSIM metrics, over a wide range of SNRs and interference levels. 
Fig. \ref{fig:cifar_snr_sinr}-(a)  highlights about $50\%$  improvement in LPIPS achieved from our scheme compared to the VAE benchmark. This can be explained due to the fact that VAEs do not maintain any `fine-grained’’ denoising mechanism to explore over probability  distributions.    
Moreover,  VAEs typically cast a standard normal distribution on the prior distribution of the latent space, which can compromise the decodability quality of semantics, while our scheme does not impose any constraint, allowing the neural encoder to obtain the semantics as expressive as it can.  
Fig. \ref{fig:cifar_snr_sinr}-(b) indicates the effect of adaptive training (over a set of CBRs provided to the neural encoder) in achieving a  robust performance during inference, with respect to changes in channel SNR and CBR values.       
Figs. \ref{fig:cifar_snr_sinr}-(c) and \ref{fig:cifar_snr_sinr}-(d)  highlight the effect SINR, indicating the dominating factor of interference in a more realistic multi-user SemCom transmission, as the LPIPS/SSIM metrics show more significant variations with changes in SINR level, than with changes in  the selected  CBR.  

Fig. \ref{fig:cifar_cbr}  compares our scheme with the matched autoencoder benchmark. In addition to the scalability issues the autoencoders have in a practical setup (discussed earlier),  our scheme showcases more than $55\%$ improvement in terms of LPIPS and around $30\%$  improvement in terms of SSIM according to Figs. \ref{fig:cifar_cbr}-(a) and \ref{fig:cifar_cbr}-(b), respectively.  Moreover, Figs. \ref{fig:cifar_cbr}-(c) and \ref{fig:cifar_cbr}-(d)   highlight the   resource-utility trade-off curve. The figures indicate  one does not need to select high CBRs when trained via adaptive CBR strategy, since  almost the same LPIPS/SSIM  can be achieved with lower CBRs (e.g., 0.2), thus saving radio resources and bandwidth.  Further visual experiments are provided in Appendix \ref{app:vis}.

\section{Conclusions} 
\vspace{-2mm}
Conditional  diffusion autoencoders have been proposed  for the nextG wireless SemCom systems.   The proposed scheme utilizes a conditional DDPM framework for the semantic decoder, and is capable of flexibly decoding semantic latents without the need for the decoder architecture to be coupled to the encoder  architecture, while also capable of learning the fine-grained details through diffusion learning.  
Theoretical analysis along with extensive simulations have been carried out to study the semantic reconstruction  performance, as well as providing insights and practical guidelines.

\section*{Acknowledgments} 
This work was supported by the Research
Council of Finland (former Academy of Finland) 6G Flagship Programme
under Grant 346208, and in part by the 6GARROW project which has received funding from the Smart Networks and Services Joint Undertaking (SNS JU) under the European Union’s Horizon Europe research and innovation programme under Grant Agreement No 101192194 and from the Institute for Information \& Communications Technology Promotion (IITP) grant funded by the Korean government (MSIT) (No. RS-2024-00435652).    
The authors wish to also acknowledge CSC - IT Center for Science, Finland, for computational resources.



\appendix

\section{Preliminaries on conditional DDPM}  \label{app:Prelim}    
Denoising diffusion probabilistic models (DDPM) are capable of learning, over iterative steps,  to reverse a  stochastic diffusion process over which the ground-truth data  is transformed into an unstructured noisy distribution.  The diffusion  process is typically defined via a Gaussian  kernel that gradually perturbs an input sample ${\vect{x}}_0 \sim p_{\vect{x}_0}(\vect{x}_0)$ following a  noise variance scheduling over $T$ steps. This can be formulated by  
\begin{equation}
    \vect x_t = \sqrt{\bar{\alpha}_t} \vect x_0 + \sqrt{\bar{\sigma}_t} \bm \epsilon, \quad \vect{x}_0 \sim p_{\vect{x}}(\vect{x}_0), \quad \bm\epsilon\sim {\cal{N}}(\bm 0, \bm I),
\end{equation}
where $\epsilon$ denotes an isotropic Gaussian, and $\alpha_t$ and $\sigma_t$ are defined according to a noise schedule with diffusion signal-to-noise ratio (SNR) of $\mathsf{SNR} = \frac{\bar{\alpha}_t}{\bar{\sigma}_t}$.  
According to the literature \cite{DM_Ho}, each step of the diffusion process is designed as expressed below, such that $\bar{\alpha}_t = \prod_{s=1}^t\alpha_s$. 
\begin{equation}\label{eq:fwd_diffusion}
    {{\vect x}}_{t} = \sqrt{\alpha_{t}} {{\vect x}}_{t-1} + \sqrt{1-\alpha_{t}} \vect \epsilon, \quad \vect{x}_0 \sim p_{\vect{x}}(\vect{x}_0), \quad  \vect\epsilon\sim {\cal{N}}(\bm 0, \bm I). 
\end{equation} 

DDPMs train a parameterized neural network $\hat{{\vect x}}_\theta(\vect x_t, \vect c; t)$ to learn the reverse of the diffusion process, where the inputs are the noisy sample $\vect x_t$, the time-step $t$, and any conditioning input,  $\vect c$, (e.g. a label, a vector, or a discrete code)  as side-information related to the ground-truth. The objective is to estimate and generate  ground-truth samples over fine-grained denoising steps. This is typically  defined as a mean squared error  (MSE) loss between 
the ground-truth and its noisy versions ${\vect x}_t$ averaged over the denoising steps $t \in [T]$: 
\begin{equation}
    \label{eqn:csm}
    \mathcal{L}(\theta) := \E_{t,(\vect x_0, \vect c), \bm \epsilon} \norm{\hat{{\vect x}}_\theta(\vect x_t, \vect c; t) - \vect x_0}_2^2.
\end{equation}

\section{Proofs} \label{app:proof}
 \textbf{Theorem 1:}   \textit{Assume $\theta^\ast_n$ be a minimizer of an $n$-sample Monte
Carlo approximation of $\mathcal{L}({\theta}, \phi)$ with respect to the decoder model $\theta$. The conditional diffusion decoder $\hat{{\vect x}}_{\theta^\ast_n}({{\vect x}}_t, \ysem; t)$ is a consistent estimator of the ground-truth, i.e., for sufficiently large number of Monte Carlo samples, we have}  
    \begin{align}
       \hat{{\vect x}}_{\theta^\ast_n}({{\vect x}}_t, \ysem; t) \overset{{P}} {\longrightarrow} \vect x_0.   
    \end{align} 

\begin{proof}
We first 
make the following assumptions:  
 \begin{itemize}
\item \textbf{Assumption 1.} Assume that  the space of parameters $\Theta$, where $\theta \in \Theta$, and the data space $\mathcal{X}$, where ${\vect x}_0 \in \mathcal{X}$, are compact.   
\item \textbf{Assumption 2.} Assume that there exists a unique $\theta^\ast \in {\Theta}$ such that $ \hat{{\vect x}}_{\theta^\ast}({{\vect x}}_t, \ysem; t) = \vect x_0$.  
\end{itemize}

 Inspecting the loss function in \eqref{eq:loss}, and using  conditional independence and the law of total expectation \cite[Ch. 7,  Proposition 5.1.]{Ross_Prob},  we can write 
    \begin{align}
         & \mathbb{E}_{\subalign{&t \sim \mathsf{Unif}[T] \\ &\vect{x}_0, \ysem \sim p(\vect{x}_0, \ysem) \\ &\vect{x}_t \sim q(\vect{x}_t | \vect{x}_0, \ysem)}}  
         \Big[ 
\norm{\hat{{\vect x}}_\theta({{\vect x}}_t, \ysem; t) - \vect x_0}_2^2
\Big]\label{eq:Diff_loss_original} \\ 
        =& \mathbb{E}_{\subalign{&t \sim \mathsf{Unif}[T]\\ &\vect{x}_0, \vect{x}_t, {\ysem} \sim p(\vect{x}_0, \vect{x}_t , \ysem )}} 
             \Big[ 
\norm{\hat{{\vect x}}_\theta({{\vect x}}_t, \ysem; t) - \vect x_0}_2^2
\Big]. \label{eq:loss_over_joint_prob}
    \end{align}
Now  define  $\nu:=(t,\vect{x}_0,\vect{x}_t,\ysem)$ and $q(\nu):=p(t,\vect{x}_0,\vect{x}_t,\ysem)$ as the joint probability distribution. Also define $f(\nu,{\theta}) := \norm{\hat{{\vect x}}_\theta({{\vect x}}_t, \ysem; t) - \vect x_0}_2^2$.  Since $t \sim \mathsf{Unif}[T]$ is independent of $(x_0, x_t ,y_\mathsf{sim}) \sim p(\vect{x}_0, \vect{x}_t , \ysem)$, \eqref{eq:loss_over_joint_prob} can be rewritten as 
    \begin{align}
       \mathcal{L}(\theta) =  \mathbb{E}_{\nu \sim q(\nu)} 
            [f(\nu,\theta)]
    \end{align}
    Therefore,  by uniform law of large numbers \cite[Lemma 2.4]{whitney1994estimation}, the Monte Carlo approximation of \eqref{eq:Diff_loss_original}, i.e.,  $\mathcal{L}^{(n)}(\theta)=\frac{1}{n}\sum_{i=1}^n f(\nu_i, \theta)$ converges uniformly in probability to $\mathcal{L}(\theta) = \mathbb{E}_{\nu \sim q(\nu)} 
    [f(\nu,\theta)]$.  

Now consider the following Lemma: 
 
\begin{lemma}[Consistency of extremum estimators] \cite[Theorem 2.1]{whitney1994estimation} \\
    \label{lemma:consistency}
    Let $\Theta$ be compact and consider a family of functions $\mathcal{L}^{(n)}: \Theta \to \mathbb{R}$. Moreover, suppose there exists a function $\mathcal{L}: \Theta \to \mathbb{R}$ such that
    \begin{itemize}
        \item $\mathcal{L}(\theta)$ is uniquely minimized at $\theta^\ast$.
        \item $\mathcal{L}(\theta)$ is continuous.
        \item $\mathcal{L}^{(n)}(\theta)$ converges uniformly in probability to $\mathcal{L}(\theta)$.
    \end{itemize}
    Then $$\theta^\ast_n := \argmin_{\theta \in \Theta} \mathcal{L}^{(n)}(\theta) \overset{P}{\to} \theta^\ast.$$
\end{lemma} 

 Let $\theta^\ast$ be the minimizer of $\mathcal{L}(\theta)$ with respect to $\theta$, then by Lemma \ref{lemma:consistency}, it indicates that $\theta^\ast_n \overset{P}{\to} \theta^\ast$, and by {Assumption 2}, it implies that  $\hat{{\vect x}}_{\theta^\ast}({{\vect x}}_t, \ysem; t) = \vect x_0$. Therefore,  $\hat{{\vect x}}_{\theta^\ast_n}({{\vect x}}_t, \ysem; t) \overset{{P}} {\longrightarrow} \vect x_0$,  and the proof is completed. 
\end{proof}

\section{Simulation Setup and neural network architectures} \label{app:neural_architecture_encoder} 
Simulations are carried out using PyTorch run on an NVIDIA V100 GPU for AI training and inference.  
Evaluations are carried out over MNIST\footnote{\url{http://yann.lecun.com/exdb/mnist}} {(resized to $32\times 32$)} and CIFAR-10\footnote{\url{https://www.cs.toronto.edu/~kriz/cifar.html}} datasets. MNIST  dataset  is split into  60,000 training
images and 10,000 testing images of hand-written digits, and CIFAR-10 dataset consists of 60000 colored images of dimension  $32 \times  32 \times 3$ pixels (height, width, channels), with separated  training and evaluation
sets containing 50000 and 10000 images, respectively. We ignore CIFAR-10 classes as our goal in this work is semantic reconstruction, not data classification.  
Evaluation metrics to measure the quality of semantic reconstruction over MNIST are  PSNR and SSIM, while for CIFAR-10, SSIM and LPIPS metrics are considered. LPIPS (learned perceptual image patch similarity) measures the perceptual similarity between images by evaluating how close the representations of the images are in high-dimensional feature space. LPIPS correlates more strongly with human perception of image quality than  metrics like PSNR and SSIM, which are more sensitive to pixel-level differences.  Lower LPIPS value indicates higher perceptual similarity between the original data and the regenerated  version at the decoder. Our plots include error bars. We have used the $\mathtt{errorbar}$ function from $\mathtt{pyplot}$ in $\mathtt{matplotlib}$   library of Python, which takes the statistical mean and standard deviation as input and plot the error bars via  $\mathtt{plt.errorbar(
        vals, mean\underline{\hspace{2mm}}vals, yerr=std\underline{\hspace{2mm}}vals)}$.    

\paragraph{Encoder's neural architecture} The encoder takes images of input size $\mathtt{channels} \times 32 \times 32$, with $\mathtt{channel}$ equal to $1$ for MNIST and $3$ for CIFAR-10 dataset respectively, and  maps them to a vector of complex channel symbols, with the number of symbols determined by the channel bandwidth ratio (CBR).  
Under MNIST dataset,  the neural encoder consists of three convolutional layers with kernel size $3 \times 3$, stride $2$, and padding $1$, having $8$, $16$, and $32$ output channels, respectively. 
Each convolutional layer is followed by a ReLU activation.  
Similarly, under CIFAR-10 dataset, the neural encoder consists of four convolutional layers 
with $64$, $128$, $256$, and $256$ output channels, respectively, 
followed by batch normalization and a ReLU activation. 
The convolutional stack progressively downsamples the input to a $32 \times 4 \times 4$ feature map over MNIST ($256 \times 4 \times 4$ over CIFAR-10), which is then flattened and projected through a fully connected layer to produce a latent vector of length $2N_c$, where $N_c$ is the number of complex symbols (with real and imaginary parts stored separately), i.e., $\mathtt{latent\underline{\hspace{2mm}}dim = 2 * int(self.input\underline{\hspace{2mm}}dim * cbr)}$.   
Finally, the latent vector is normalized to have unit $\ell_2$-norm to satisfy the average power constraint.   

For the benchmarks, the encoder  architecture follow the same architecture as  the one employed for our scheme, and the decoder counterpart mirrors the encoding functionality for the autoencoder benchmark (matched decoder). While for the VAE benchmark, flattened features are further passed through two parallel linear layers which respectively estimate the mean $\mu$ and log-variance  $\log \sigma^2$  of the posterior distribution over latent variables. VAE sampling is then performed using the standard reparameterization trick, where $\mathbf{z} = \boldsymbol{\mu} + \boldsymbol{\sigma} \odot \boldsymbol{\epsilon}$, $\boldsymbol{\epsilon} \sim \mathcal{N}(0, I).$
 A normalization step ensures the power normalization for the sampled latent vector before transmission, emulating practical communication constraints.  
 VAE training is carried out with the Adam optimizer (learning rate 0.001), jointly training the encoder and decoder parameters with a loss function that combines (with equal weights) a pixel-wise MSE reconstruction loss with the KL divergence regularizer, ensuring the latent space follows a Gaussian prior.  

\paragraph{Diffusion decoder architecture}  
The implementation of our diffusion model is inherited from  the seminal DDPM framework  \cite{DM_Ho}\footnote{\url{https://github.com/hojonathanho/diffusion}} implemented it in PyTorch (the original code was in TensorFlow).   
We further incorporate the conditioning mechanism into this framework, where the conditioning mechanism is developed  upon  \cite{cdiff}\footnote{\url{https://github.com/neillu23/CDiffuSE}} and adopted to the image generation task (the original code is for speech signal generation).    

The core learning model as the backbone of our decoder is a time-conditioned U-Net architecture, integrated into a SemCom data pipeline, and optimized to operate jointly with neural encoders.   
The U-Net uses a base spatial feature dimension $\mathtt{dim=32}$ with four resolution stages determined by the multiplier tuple $(1, 2, 4, 8)$. 
The input to the U-Net is the diffused original samples ${{\bm x}}_t$ concatenated with the conditioning input (noisy semantic latent vector),  where the semantics  are (padded and) reshaped into a $(\mathtt{channels}, 32, 32)$ tensor to match the image spatial dimensions, ensuring that semantic conditioning  information can be injected directly into the spatial feature maps during training and denoising.   
The network begins with an initial convolution that takes two-channel input (original channels concatenated with the conditional tensor) and maps it to an initial dimension $\mathtt{init\underline{\hspace{2mm}}dim = (dim // 3) * 2}$
From there,  each stage of the encoder consists of two ConvNeXt-style residual blocks \cite{ConvNext} with depthwise convolution of kernel size 7,  
followed by two pointwise convolutions, GELU activation, and group normalization (with a single group)
with learnable skip connections. 
ConvNeXt blocks are extended with a conditioning pathway that  incorporates time-step embeddings (from the diffusion process)  through a sinusoidal position embedding projected by an MLP to dimension of 
$\texttt{time\underline{\hspace{2mm}}dim = dim * 4 = 128}$, which is added to the block's activation. 
This enables the network to adapt its representation across different noise levels in the generative process.
Within each resolution stage, a linear attention module (single-head dot-product attention with linear complexity) is wrapped in a pre-norm residual structure, i.e., the feature maps are  normalized, and processed by the attention operation, and finally the original input is added back to the attention output.
Spatial resolution is reduced between U-Net stages across the encoder path  by a downsampling operation via  a regular convolution layer (kernel size $4$, stride $2$, padding $1$), 
enabling the network to progressively increase its receptive field and capture higher-level semantic features until the bottleneck is reached.  At the bottleneck, two ConvNeXt blocks are applied with a full multi-head attention layer in between, capturing both local and global context at the coarsest spatial resolution.
The decoder path mirrors the encoder in reverse order. Each upsampling stage concatenates the skip connection from its symmetric encoder stage, then processes the tensors with two ConvNeXt blocks, a linear attention layer, and an upsampling layer. The final stage employs a  ConvNeXt block followed by a $1 \times 1$ convolution,    
defaulting back to the original channel dimensions    $\mathtt{out\underline{\hspace{2mm}}dim} = 1$ for MNIST and $\mathtt{out\underline{\hspace{2mm}}dim} = 3$ for CIFAR-10, respectively, so as to having output  samples with the same spatial dimension  as the  input tensor.   

\section{Training details}  \label{app:train}
Training of our scheme is carried out in two different  scenarios: 
i)  Fixed training  (for 10 epochs over MNIST and 50 epochs over CIFAR-10 dataset), where  a fixed CBR of $0.3$ over MNIST and $0.4$ over CIFAR-10 is set for the neural encoder, and the joint training of the neural encoder and diffusion decoder is done for a fixed CBR. Nevertheless, the evaluations are still carried out under different CBRs to highlight the scalability  of our approach, i.e, no need for re-training the pipeline for different CBRs. Rather,  once trained with an arbitrary neural encoder (mimicked here as a neural encoder with CBR of 0.3 or 0.4), the CDiff decoder is inherently capable of adapting to different neural encoders of different types/sizes via the padding-and-masking condition mechanism. In the fixed training scenario, to test how well the model generalizes to lossy links with CBRs lower than the training CBR (or scenarios where less radio resources are available in inference compared to the training stage), stochastic masking (aka soft masking) is applied to the neural encoder's output (followed by the re-normalization to preserve average transmit power constraint) before sending the semantic latents over the wireless channel.  In stochastic masking, we randomly drop or scale down parts of the neural encoder output probabilistically (instead of hard-zeroing as in hard masking). Intuitively, this basically  allows us to  simulate a channel scenario where each symbol has a chance to get through independently. 
ii) Adaptive training  (for 20 epochs over MNIST and 50 epochs over CIFAR-10 dataset), where the neural encoder is provided with a selectable set of channel bandwidth ratio (CBR) from a predefined set $\{ \texttt{CBR1}, \texttt{CBR2}, \dots \}$. 
The feature extraction stage is shared across all CBR settings and the same neural architecture is assumed, where for each CBR value, a dedicated fully connected projection layer maps this shared feature vector to a latent vector of length $2N_c$, where $N_c$ is the number of complex symbols for the selected CBR.  We defined the set of CBRs for the adaptive neural encoder as $\mathtt{cbr\underline{\hspace{2mm}}list=[0.2, 0.25, 0.3, 0.35, 0.4, 0.45]}$, and the joint training is done via randomly choosing from this set in each epoch.

We set the number of denoising steps to 200, and use  adaptive moment estimation (Adam) optimizer with  learning rate  $\lambda = 10^{-3}$.  To stabilize the training process,  exponential moving average (EMA) method  is  implemented \cite{EMA} {for model update in each epoch,  where a copy of the previous weights is kept, and the weighted average  between the previous and new version of the weights are calculated as the model update. 
 For the conditioning weights $0<w_t<1$, we consider a linear scheduling.  For the variance scheduling $\beta_t$, we also consider a linear scheduling, starting from $\beta_0 = 0.0001$ to $\beta_T = 0.0095$.  The  values were simply obtained out of trial-and-error.  
The SNR of communication links is  defined as:
	\begin{equation}
		\Gamma_{}=10\log_{10}\frac{P}{\sigma_{}^2}~ \mathrm{~(dB)}. 
		\label{eq:snr}
	\end{equation}
This basically represents the ratio of the average power of the encoded data-stream of the latent space (i.e., the channel input) to the average AWGN noise power. 
	Without loss of generality, we set the average signal power  to $P=1$ for all experiments, while  varying  the SNR by setting the noise variance $\sigma_{}^2$.    Training SNR is chosen uniformly from  the range of $-10 \leq \Gamma_{}^{\mathtt{train}} \leq 10$ dB during training, while the evaluations are carried out over a wider range $-10 \leq \Gamma_{}^{\mathtt{test}} \leq 30$ dB to highlight the generalization of our scheme.

\section{Further experiments}\label{app:vis}
Figs. \ref{fig:vis_mnist}, \ref{fig:vis_mnist_interf}, \ref{fig:vis_cifar}, and \ref{fig:vis_cifar_interf} further provide the visualizations on the reconstructed samples at our CDiff decoder, where the performance metrics are shown below each sample.  

\begin{figure*}[t]
\centering
    \begin{subfigure}[]{0.48\textwidth}
        \centering  
        \vspace{-0.1mm}
        \includegraphics[width=\linewidth, trim={0.0in 0.0in 0.0in  0.0in},clip]{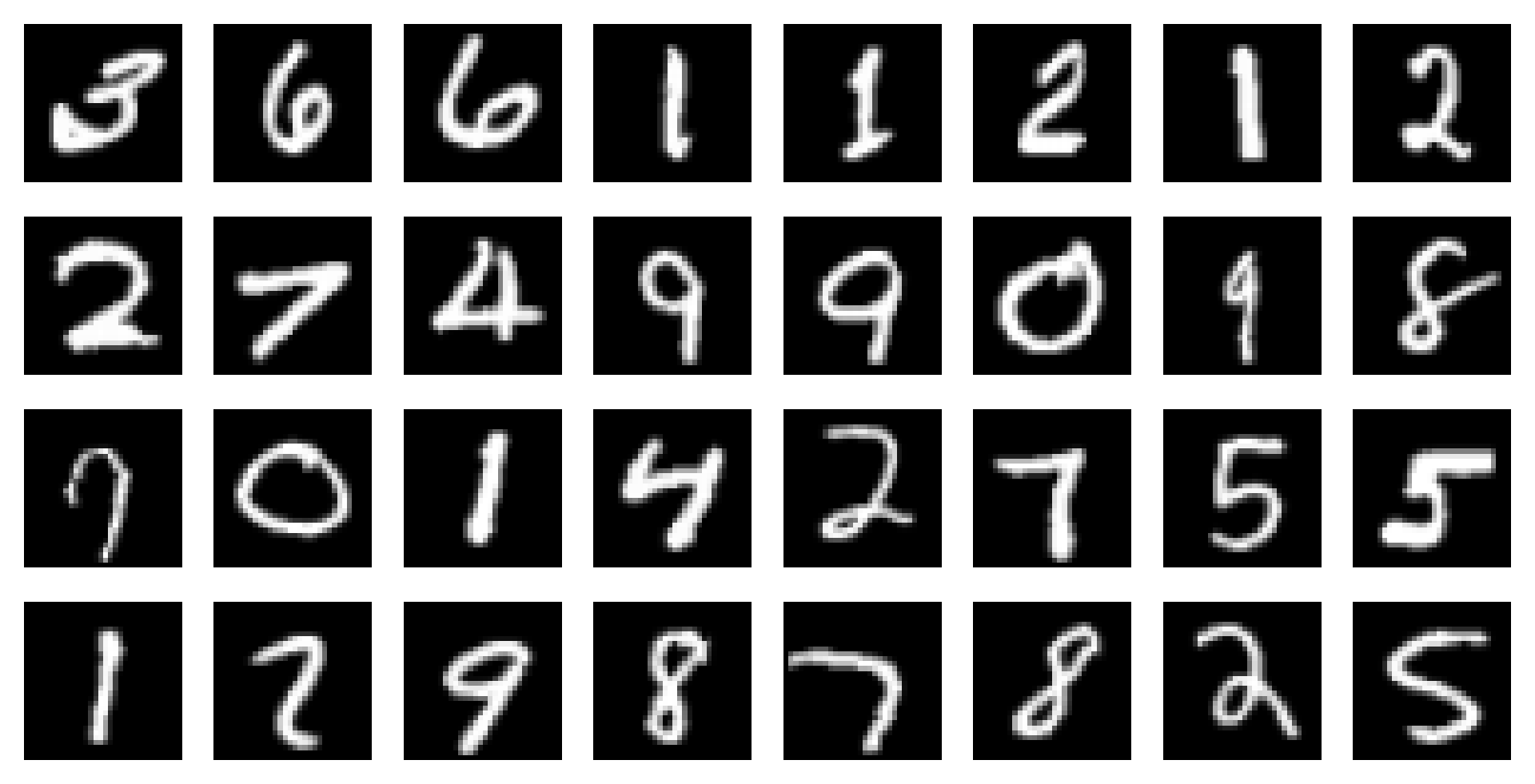}
        \caption{Original samples}
    \end{subfigure}
     \begin{subfigure}[]{0.48\textwidth}
        \centering  
        \vspace{0.1mm}
        \includegraphics[width=\linewidth, trim={0.0in 0.0in 0.0in  0.0in},clip]{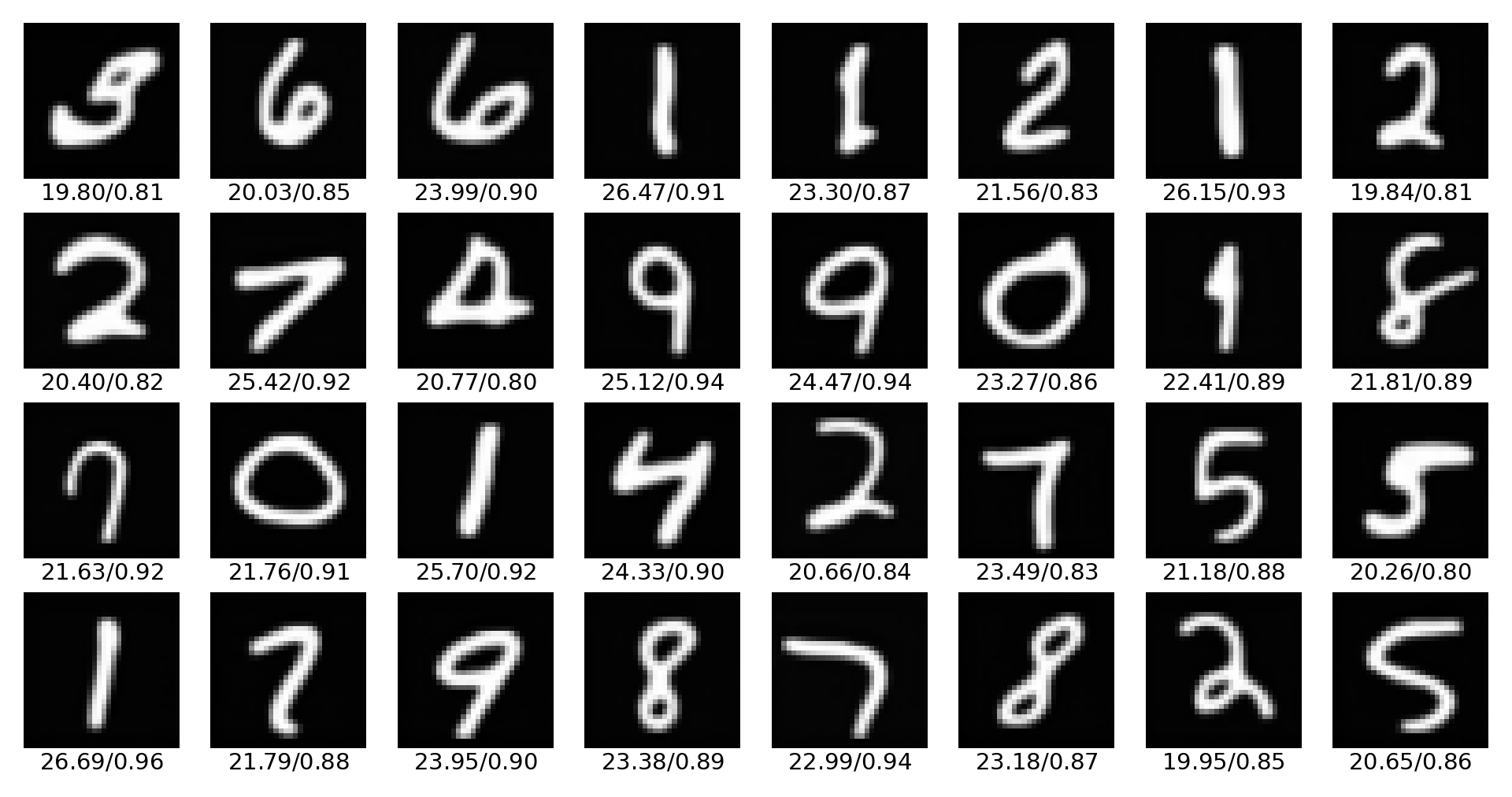}
        \caption{Reconstructed samples} 
    \end{subfigure}
    \caption{Data visualization over MNIST for test SNR 0 dB. PSNR/SSIM values are shown below each sample.    
    } 
    \label{fig:vis_mnist}
    \vspace{0mm}
\end{figure*}

\begin{figure*}[h!]
\centering
    \begin{subfigure}[]{0.48\textwidth}
        \centering  
        \vspace{-0.1mm}
        \includegraphics[width=\linewidth, trim={0.0in 0.0in 0.0in  0.0in},clip]{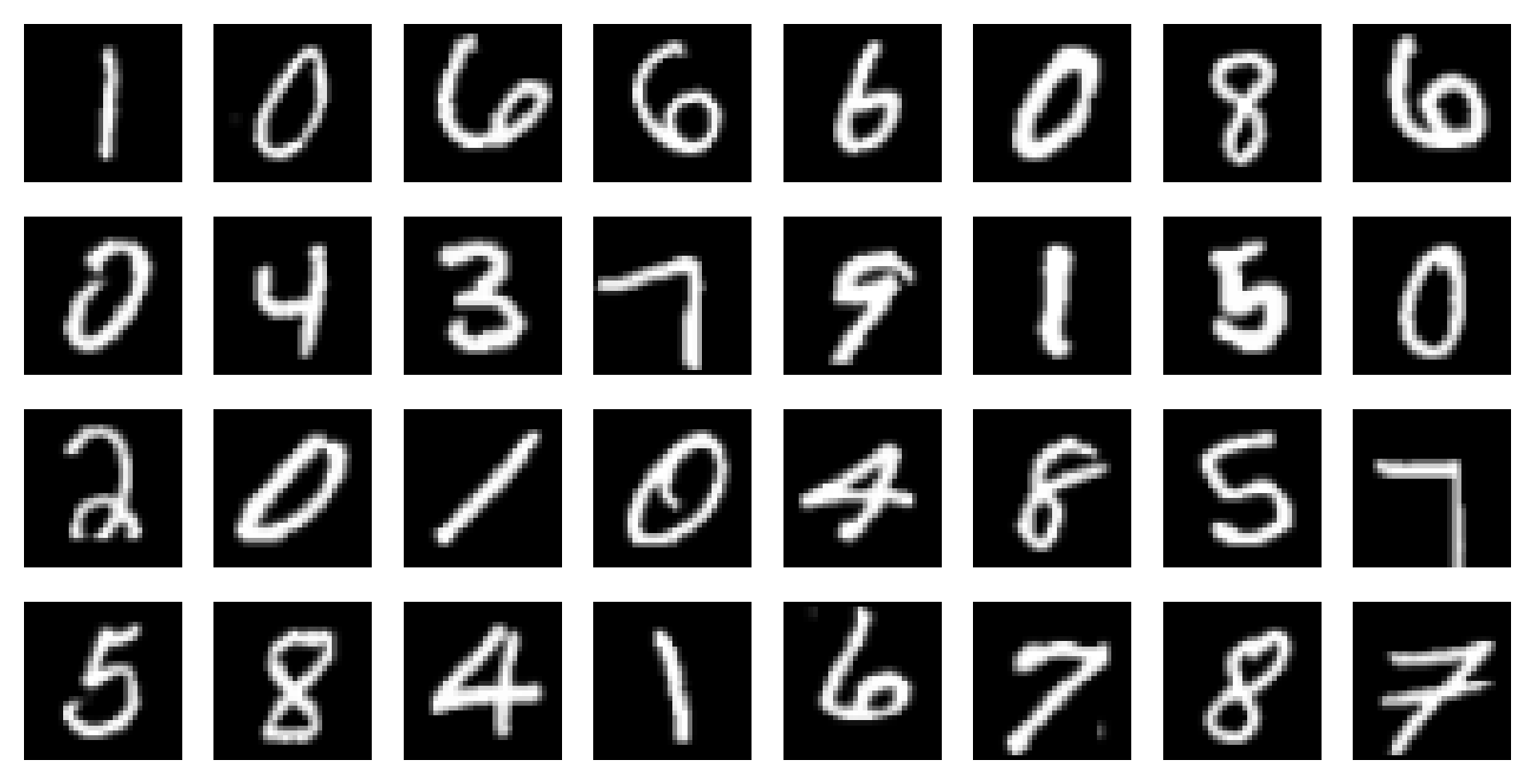}
        \caption{Original samples}
    \end{subfigure}
     \begin{subfigure}[]{0.48\textwidth}
        \centering  
        \vspace{0.1mm}
        \includegraphics[width=\linewidth, trim={0.0in 0.0in 0.0in  0.0in},clip]{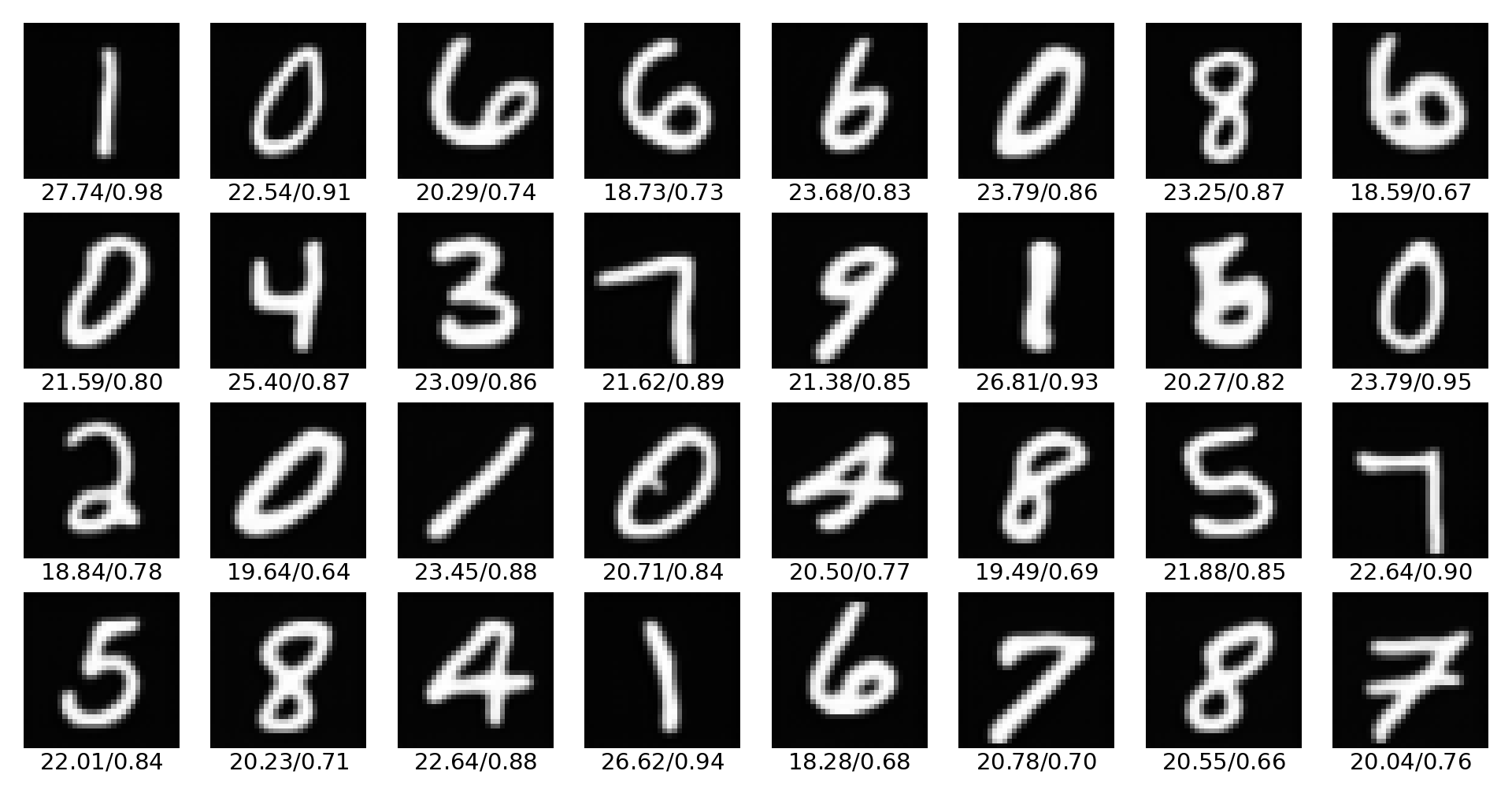}
        \caption{Reconstructed samples} 
    \end{subfigure}
    \caption{Data visualization for multi-user scenario  over MNIST. A convex combination with coefficients  of 0.8 and 0.2 are assumed for the source semantics and the interfering  latents (equivalent $\texttt{SINR} \approx -2.1$ dB).   Test SNR is set to 0 dB. PSNR/SSIM values are shown below each sample.    
    } 
    \label{fig:vis_mnist_interf}
    \vspace{0mm}
\end{figure*}

\begin{figure*}[t]
\centering
    \begin{subfigure}[]{0.48\textwidth}
        \centering  
        \vspace{-0.1mm}
        \includegraphics[width=\linewidth, trim={0.0in 0.0in 0.0in  0.0in},clip]{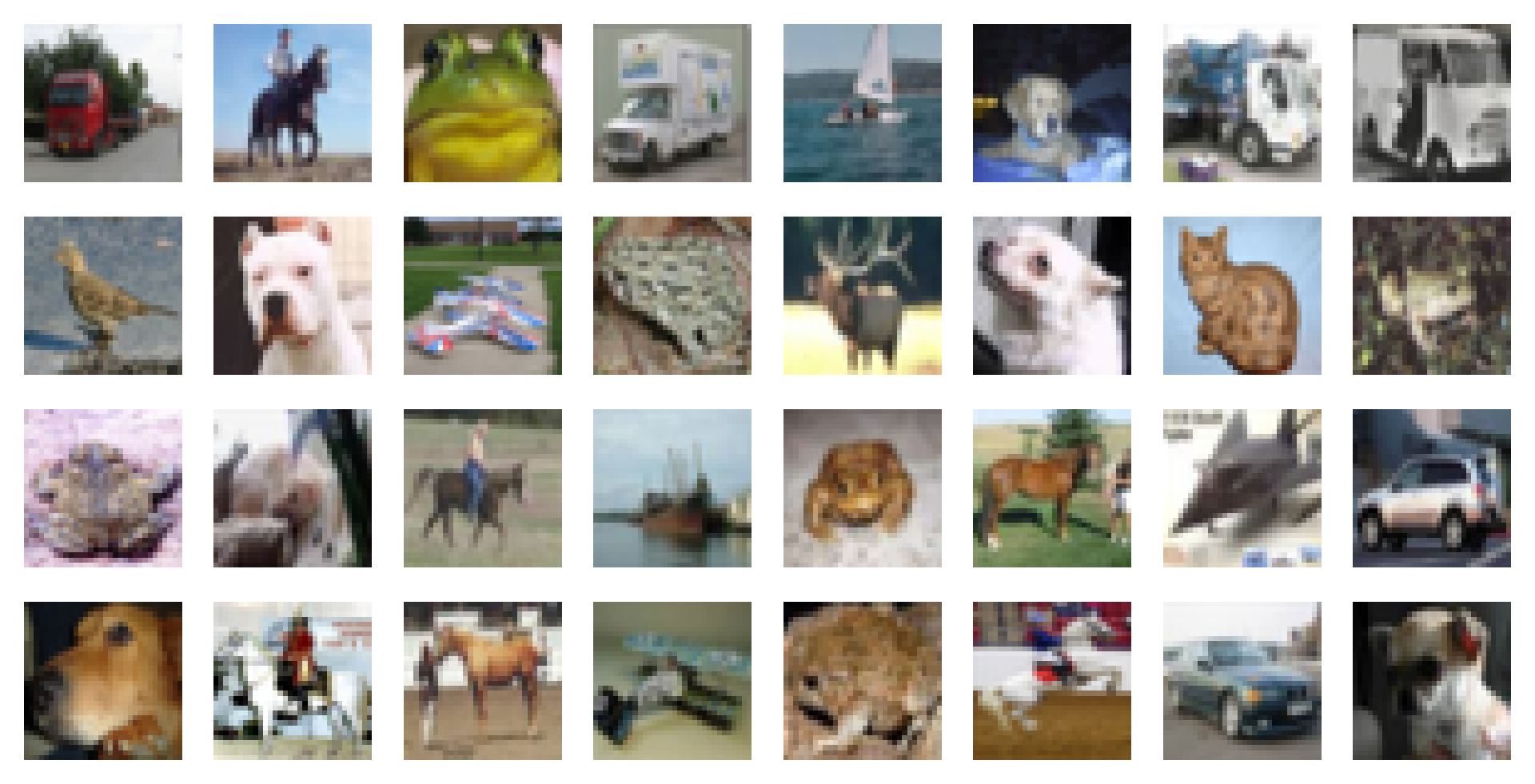}
        \caption{Original samples}
    \end{subfigure}
     \begin{subfigure}[]{0.48\textwidth}
        \centering  
        \vspace{0.1mm}
        \includegraphics[width=\linewidth, trim={0.0in 0.0in 0.0in  0.0in},clip]{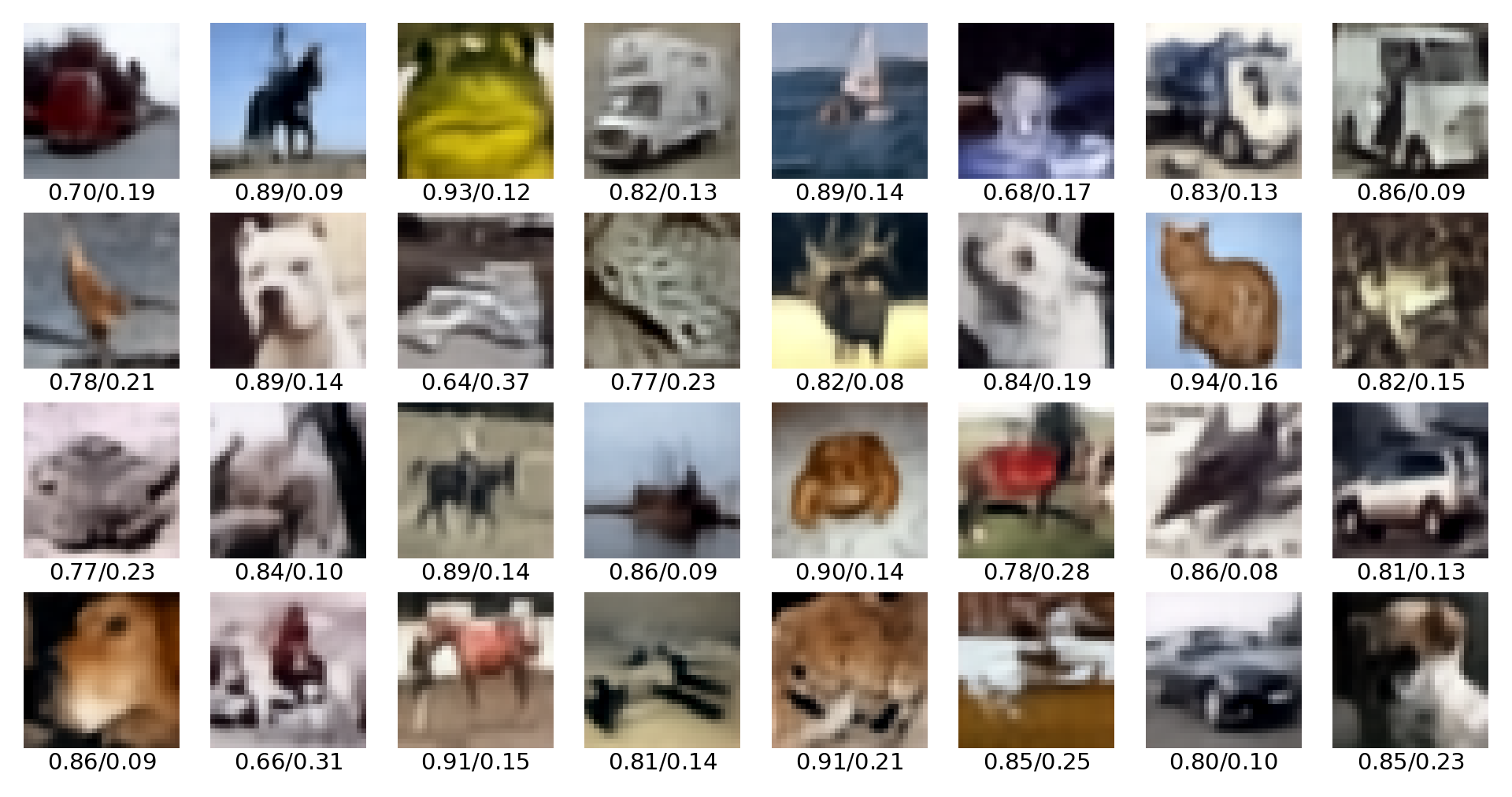}
        \caption{Reconstructed samples} 
    \end{subfigure}
    \caption{Data visualization over CIFAR-10 for test SNR 10 dB. SSIM/LPIPS values are shown below each sample.    
    } 
    \label{fig:vis_cifar}
    \vspace{0mm}
\end{figure*}

\begin{figure*}[tbph!]
\centering
    \begin{subfigure}[]{0.48\textwidth}
        \centering  
        \vspace{-0.1mm}
        \includegraphics[width=\linewidth, trim={0.0in 0.0in 0.0in  0.0in},clip]{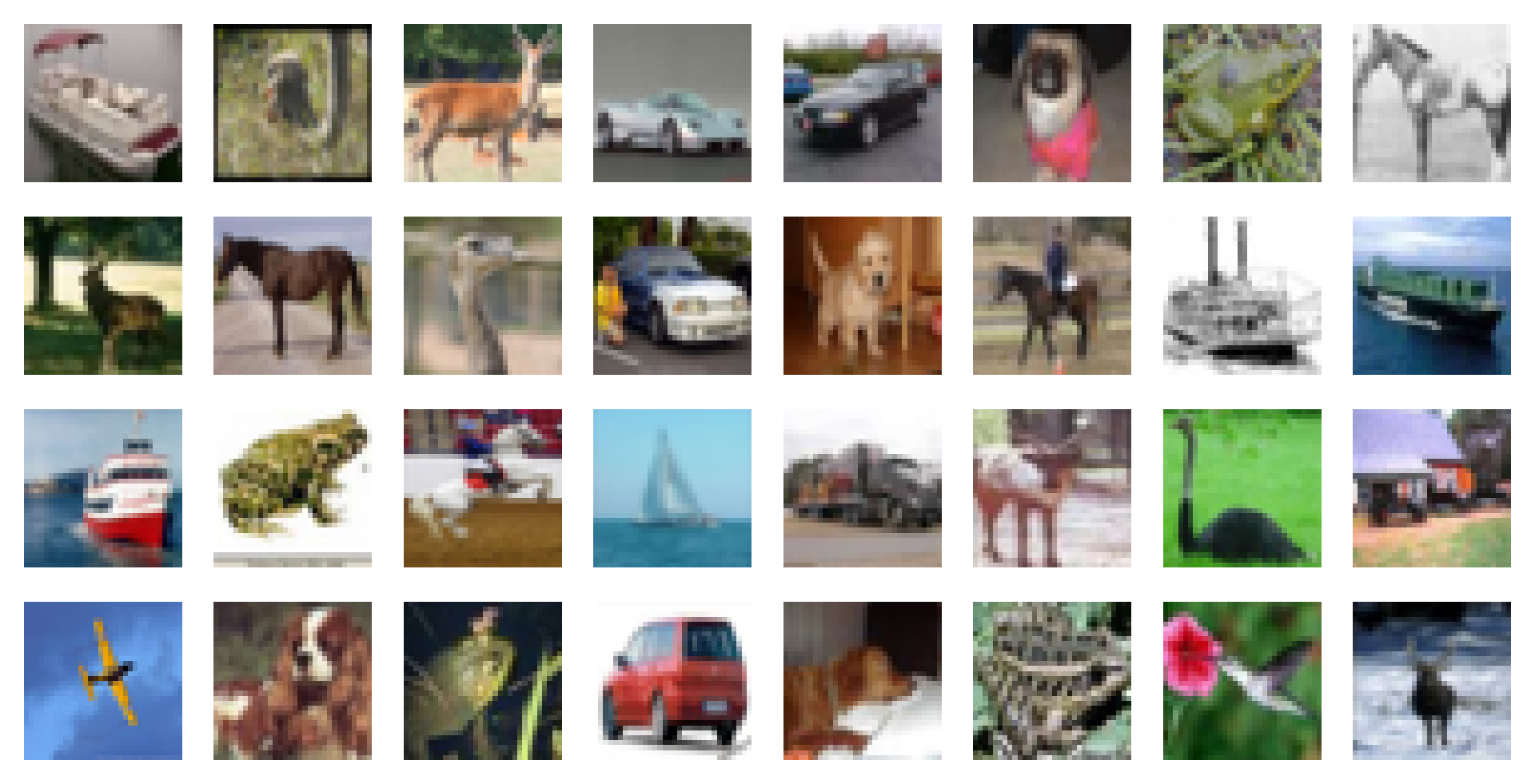}
        \caption{Original samples}
    \end{subfigure}
     \begin{subfigure}[]{0.48\textwidth}
        \centering  
        \vspace{0.1mm}
        \includegraphics[width=\linewidth, trim={0.0in 0.0in 0.0in  0.0in},clip]{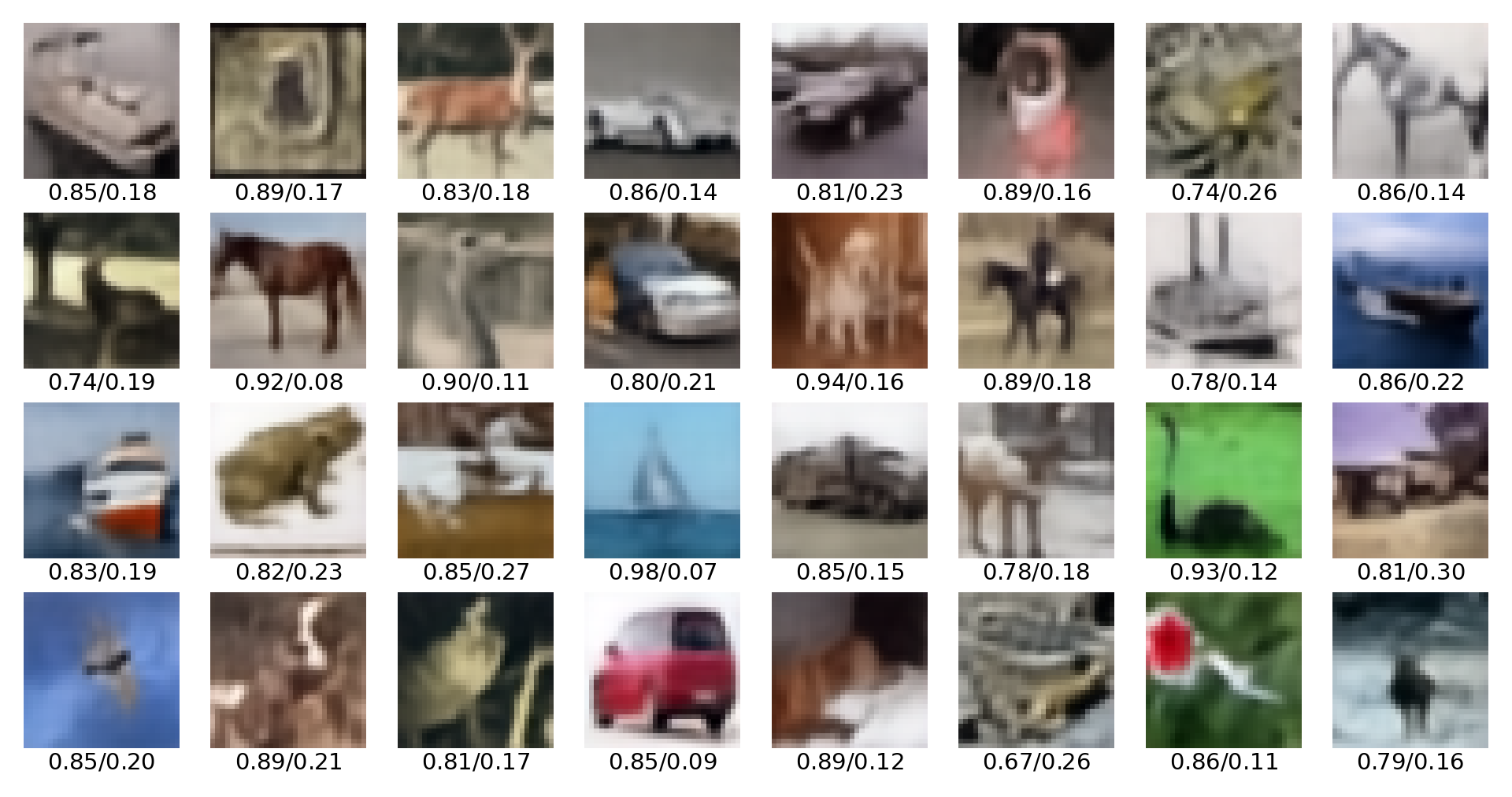}
        \caption{Reconstructed samples} 
    \end{subfigure}
    \caption{Data visualization for SemCom reconstitution under multi-user interference over CIFAR-10. A convex combination with coefficients  0.9  and 0.1 are assumed for the source semantics and the interfering  latents respectively (equivalent $\texttt{SINR} \approx 16$ dB).   Test SNR is set to 20 dB. SSIM/LPIPS values are shown below each sample.   
    } 
    \label{fig:vis_cifar_interf}
    \vspace{0mm}
\end{figure*}

\section{Limitations}  \label{app:limits}
We would like to point out the limitations of our work, in terms of the assumptions, datasets used, computation efficiency, and practical implications.  

\paragraph{Performance under non-stationary fading channels}  While in this work, we modeled the channel as an additive Gaussian noise (AWGN channel), the performance of SemCom systems under non-stationary wireless channel models needs careful consideration. One approach to solve this issue would be to re-train or fine-tune the systems (joint end-to-end fine-tuning) over \emph{real-time} wireless channels. However, another  challenge arises here: the channel coherence time of wireless channels in practice  are relatively larger than the
rate at which data samples are processed for training.
Hence, only a few channel realizations might be observed over
every batch, which may not be favorable in terms of generalization to a wide range of channel realizations. 
Another simple approach might be to rely on traditional channel equalizers before feeding the semantic latents to the decoder.  

\paragraph{Out-of-distribution (OOD) performance} 
OOD  performance limitation of our scheme should be revisited with respect to both the wireless channel models (discussed above), and also  the datasets over which the models are trained.  SemCom systems so far have shown to be typically dataset-dependent. This can be seen as one major challenge with such systems in practical deployments. Still, the fact that datasets are shared among the communication parties can be seen as a reasonable interpretation: In the context of SemCom, the transmitter and receiver share a common \emph{knowledge base} (KB) to realize a common ``mutual understanding'' of the semantic meanings. This is typically  realized as a dataset shared between both sides. In practice, transmitter (e.g., the terminal device equipped with neural encoder) can signal the dataset identifier (ID) to the receiver (e.g., the cellular gNodeB equipped with a foundation decoder model), and the receiver can activate its corresponding model (among pool of models)  pre-trained on the corresponding dateset ID; and then the communication gets started after handshaking on the KB.

\paragraph{Sampling speed and computation efficiency} A practical limitation of our work is that DDPMs may not be computation efficient  due to their iterative denoising. Nevertheless,  there is an active line of research within the AI community, exploring  ways to accelerate the generation process, and we envision that more computation-efficient diffusion models will evolve in the near future. 
Moreover, this can be viewed in another way: SemCom framework  makes the communication systems radio resource-efficient, saving the transmit power, physical resource blocks, and also the  bandwidth, due to sending only the minimal relevant information rather than the entire datastream.  Notably, this is achieved by moving the processings from PHY to AI. The AI processing can be done at cloud, over-the-top (OTT) server from device perspective, or core network from gNodeB perspective, in a practical nextG cellular system, thanks to richer compute resources available in those network entities. Hence,  from  communication perspective, Semcom is actually making the systems more efficient, particularly  for resource-constrained wireless systems.  

\paragraph{Definitions, specification, and standardizations}
Concepts and definitions we used in this work for semantic latents, conditioning mechanism, neural encoding/decoding functionality, etc.,  may not be the same as the ones which will be specified in practical  SemCom deployments in nextG.    
For a wide adoption of SemCom, proper definitions for  semantic information, semantic noise, semantic interference, semantic error rate, knowledge base, etc., should be specified and agreed upon cellular partners to be standardized with respect to signaling interfaces, data, and control information elements to ensure interoperability in real-world setups.

\end{document}